\documentclass[5p]{elsarticle}
\pdfoutput=1
\usepackage{amsmath,amssymb,amsfonts}
\usepackage[ruled,vlined]{algorithm2e}
\usepackage{algorithmic}
\usepackage{graphicx}
\usepackage{textcomp}
\usepackage{xcolor}
\usepackage{booktabs}
\usepackage{url}
\usepackage{ragged2e}

\journal{Journal of \LaTeX\ Templates}

\newcommand{\E}{\mathcal{E}}

\newcommand{\lvari}{\tilde{\mathcal{V_i}}}
\newcommand{\vari}{\mathcal{V}_i}
\newcommand{\fset}{A_j^{\vari}}

\begin{document}

\begin{frontmatter}
\title{Combining Embeddings and Fuzzy Time Series for  High-Dimensional Time Series Forecasting in Internet of Energy Applications}


\author[ppgee,minds]{Hugo Vinicius Bitencourt}
\ead{hugovynicius@ufmg.br}
\author[minds]{Luiz Augusto Facury de Souza}
\author[minds]{Matheus Cascalho dos Santos}
\author[ifnmg]{Petrônio Cândido de Lima e Silva}
\ead{petronio.candido@ifnmg.edu.br}
\author[minds]{Frederico Gadelha Guimarães\corref{mycorrespondingauthor}}
\cortext[mycorrespondingauthor]{Corresponding author}
\ead[url]{https://minds.eng.ufmg.br/}
\ead{fredericoguimaraes@ufmg.br}

\address[ppgee]{Graduate Program in Electrical Engineering, Universidade Federal de Minas Gerais, Belo Horizonte, Brazil}
\address[minds]{Machine Intelligence and Data Science ({MINDS}) Laboratory, Federal University of Minas Gerais, Belo Horizonte, Brazil}
\address[ifnmg]{Federal Institute of Education Science and Technology of Northern Minas Gerais, Januária Campus, Brazil}




\begin{abstract}
The prediction of residential power usage is essential in assisting a smart grid to manage and preserve energy to ensure efficient use. An accurate energy forecasting at the customer level will reflect directly into efficiency improvements across the power grid system, however forecasting building energy use is a complex task due to many influencing factors, such as meteorological and occupancy patterns. In addiction, high-dimensional time series increasingly arise in the Internet of Energy (IoE), given the emergence of multi-sensor environments and the two way communication between energy consumers and the smart grid. Therefore, methods that are capable of computing high-dimensional time series are of great value in smart building and IoE applications. Fuzzy Time Series (FTS) models stand out as data-driven non-parametric models of easy implementation and high accuracy. Unfortunately, the existing FTS models can be unfeasible if all features were used to train the model. We present a new methodology for handling high-dimensional time series, by projecting the original high-dimensional data into a low dimensional embedding space and using multivariate FTS approach in this low dimensional representation. Combining these techniques enables a better representation of the complex content of multivariate time series and more accurate forecasts.

\end{abstract}


\end{frontmatter}


\section{Introduction}
\label{introduction}

We are on the cusp of a new age of ubiquitous networks (i.e. networks available anywhere, anytime) and communication that has transformed corporate, community and personal spheres. Connections are multiplying, creating an entirely new dynamic network - the Internet of Things (IoT). Internet of Things means a global network of interconnected objects (e.g. RFID tags, sensors, actuators, mobile phones) that are uniquely addressable and based on standard communication protocols \cite{Union2015IoTReport} \cite{Gubbi2013IoTSurvey} \cite{Miorandi2012IoTSurvey}.


The Internet of Things may impact various aspects of everyday life and the behavior of potential users, and popular demand combined with technological advances could drive the widespread diffusion of an IoT that could contribute to economic development. Sensor nodes and actuators distributed throughout homes and offices can make our lives more comfortable, for example: room heating can be adjusted to our preferences and the weather; domestic incidents can be avoided with appropriate monitoring and alarm systems; and energy can be saved by automatically turning off electrical appliances when they are not needed. Roads can be equipped with tags and sensor nodes that send important information to traffic control centers and transportation vehicles to better direct traffic. Sensors can also be used to monitor water quality, diagnose and treat diseases \cite{Reka2018FutureEffectualIoT} \cite{Yasir2020IoE} \cite{Michele2016IoTSurvey}.


The integration of sensor nodes, actuators, smart meters and other components of the electrical grid together with information and communication technology is called the Internet of Energy (IoE). IoE uses the bidirectional flow of energy and information within the electrical grid to gain deep insights into electricity usage and benefit from future actions to increase energy efficiency. IoE is a subcategory of IoT and an indispensable component in the implementation of smart grids \cite{Manar2015IoE} \cite{Yasir2020IoE} \cite{Shahinzadeh2019IoE}.


Smart grids can be defined as the modernization of electrical power systems to achieve a fully automated power grid by integrating all connected users. In smart grids, all players (suppliers and consumers) are digitally connected. Smart grid technology promises to make power systems safer, more reliable, more efficient, more flexible and more sustainable. It achieves these goals through the use of IoE technology in power grids \cite{Jaradat2015IotSmartGrid}.


Smart buildings or smart homes use IoE devices to monitor various building characteristics, analyze the data, and generate insights into usage patterns and trends that can be used to optimize the building environment and operations. The concept of smart homes can be easily extended to all types of buildings such as industrial, commercial and residential buildings \cite{Lobaccaro2016SurveySmartHomes}.


In smart buildings, all digital devices are interconnected to form an IoE network that automates  or assists users in various ways, for example, IoE devices monitor home energy consumption and users can control their home electricity usage through bidirectional communication with home appliances \cite{Shahinzadeh2019IoE}.


There is a need to add computational intelligence at all levels of the smart grid to mitigate potential uncertainties in the electrical grid. An important feature of smart buildings is the ability to forecast energy consumption, as utilities are expected to learn the patterns of energy consumption and predict demand to make appropriate decisions. A smart learning mechanism enables energy planning strategies that contribute to more efficient use of electricity, higher end-use efficiency, and effective use of electricity infrastructure \cite{Manic2016IntelligentBuildings} \cite{Hammami2020ANNOnlineSmartGrids} \cite{Mohammadi2018SurveyDPIoTBigData}. 


The ability to control and improve energy efficiency and predict future demand is imperative for smart buildings. Accurate energy forecasting at the customer level will reflect directly into efficiency improvements across the power grid system, but forecasting building energy use is a complex task due to many influencing factors, such as meteorological and occupancy patterns.


In the context of IoE, an enormous amount of sensor nodes collect sensory data over time for a variety of applications, resulting in a huge amount of continuous data and Big Data streams. The data is continuously recorded from various data sources and each sensor generates a streaming time series, where each dimension represents the measurements recorded by a sensor node, resulting in a high-dimensional time series. Formally, an IoT application with $M$ sensors generates an $M$-dimensional time series. IoT and IoE are one of the most important sources of Big Data, as they rely on connecting a large number of devices to the Internet to continuously report the state of  the environment.


In addition, IoE devices may suffer from unavoidable aging effects or faults in their embedded electronics, then many such data may exhibit errors and noise during acquisition and transmission. Furthermore, the physical phenomena to be monitored may also change over time due to seasonality.


In this sense, forecasting methods for high-dimensional time series have become one of the active topics in machine learning researches. Methods capable of handling high-dimensional time series are of great value in IoE applications. The analysis of such datasets presents significant challenges, both from a statistical and numerical perspective.


Fuzzy Time Series (FTS) methods became attractive due to their ease of implementation, low computational cost, predictive accuracy, and model interpretability. However, as the dimensionality of the time series increases, FTS methods significantly lose accuracy and simplicity. Since each variable has its own fuzzy sets and the number of rules in a multivariate FTS model is given by a combination of the fuzzy sets, the number of rules can grow exponentially with the number of variables \cite{Guimaraes2020HyperOtimFTS} \cite{Severiano2020NSFTS}. Therefore, it is neither practical nor computationally sound to impose all input data on an FTS forecasting model. The challenge then is to design an optimal set of inputs based on their strong correlations with the target output.


To overcome these challenges presented above, we introduce here a new methodology for forecasting high-di\-men\-sional time series called $\gamma$FTS (Embedding Fuzzy Time Series). We apply a data embedding transformation and use FTS models in a low dimensional, learned continuous representation. Embedding allows us to extract a new feature space that better represents the complex content of multivariate time series data for the subsequent forecasting task. The combination of these techniques enables better representation of the complex content of high-dimensional time series and accurate forecasting.


The $\gamma$FTS was tested to solve the problem of energy consumption forecasting of devices in smart buildings. The methodology was used to extract a new feature space that better represents the content of the multivariate time series of energy consumption for the subsequent forecasting task. The embedding methods allow us to extract the relevant information that supports the prediction of the target variable.


The main contributions of our work are:

\begin{enumerate} 
    \item Investigating the potential benefits of a method that combines an embedding transformation and a fuzzy time series forecasting approach for dealing with high-dimensional time series data; 
    \item We present a new methodology that has great value for IoE applications in smart buildings and can help homeowners reduce their electricity consumption and provide better energy saving strategies; 
    \item The proposed methodology generates models that are fully readable, transparent, testable and explainable.
\end{enumerate}



The remainder of the work is organized as follows. In section~\ref{literature_review} related work, both from the standpoint of application and methodology, is presented. In section~\ref{efts} the proposed approach is described in detail. Section~\ref{experiments} describes the case studies of three smart building applications used to test our method. The results of the case studies are presented and discussed in section~\ref{results}. Section~\ref{conclusions} concludes the paper.
    

\section{Literature Review}
\label{literature_review}

In this section, we present the related works, both from the point of view of application and methodology. This section describes Wireless Sensor Networks, presents current methods for energy consumption forecasting in Smart Buildings and defines fuzzy time series and embedding transformation.


\subsection{Energy Consumption Forecasting}
\label{energy_consumption}

Wireless Sensor Networks (WSN) is a key component of the Internet of Energy and the Internet of Things. WSNs consist of many devices called sensor nodes. Sensor nodes are autonomous, compact, low-power sensors integrated with a low-power embedded CPU, memory, and wireless communication, and may also have local data processing and multi-hop communication \cite{Zheng2015WSN} \cite{Vieira2003WSNSurvey}.


Sensor nodes can be used to detect or monitor a variety of physical parameters or conditions (e.g., humidity, temperature). WSNs offer numerous possibilities for power grids, such as power monitoring, demand-side energy management, coordination of distributed storage, and integration of renewable energy generators.


Wireless Sensor Networks have been used to collect data to analyze the behavior and proper use of energy. Prediction of energy consumption is very important for smart buildings as it helps to reduce power consumption and achieve better energy and cost savings. Forecasting energy consumption allows home and building managers to plan energy use over time, shift energy use to off-peak times, identify goals for energy savings, and make more favorable plans for energy purchases \cite{Manic2016IntelligentBuildings} \cite{Mocanu2016DPBuildingEnergyConsumption}.


Smart Buildings is one of the most popular IoE applications that can help increase the energy efficiency of the home. Home appliances can be connected to the internet to help homeowners have better control over their home utility and spending, and can be used in conjunction with meteorological data to predict appliance energy consumption \cite{Mohammadi2018SurveyDPIoTBigData} \cite{Manic2016IntelligentBuildings}.


Forecasting energy consumption in smart buildings is a time series problem consisting of one- or high-dimensional features. These time series usually have seasonal variations and irregular trend components. The data recorded by sensor nodes may also contain missing values, outliers, and uncertainties. Therefore, accurate energy consumption forecasting is a difficult task due to these unpredictable disturbances and noisy data.


Several machine learning models have been used to predict energy consumption using historical data collected from sensor nodes. These techniques have been developed to improve the quality of the power grid and optimise energy utilisation.


Candanedo et al. \cite{Candanedo2017Appliances} implemented and evaluated four data-driven predictive models for appliance energy consumption in a low-energy house in Belgium. The authors tested Multiple Linear Regression (MLR), Support Vector Machine with Radial Kernel (SVM-radial), Random Forest (RF), and Gradient Boosting Machines (GBM). The best model was GBM, which could explain 57\% of the variance ($R^2$) and achieved an RMSE of 66.65, a MAE of 35.22 and a MAPE of 38.29\% in the testing set when all features were used.


Chammas et al. \cite{Chammas2019AppliancesMLP} proposed a Multilayer Perceptron (MLP) with four hidden layers and 512 neurons in each layer to predict the energy consumption of appliances. The MLP model was able to predict 56\% of variance with 66.29 of RMSE, 29.55 MAE and 27.96\% MAPE in the testing set when all features were used.


Mocanu et al. \cite{Mocanu2016DPBuildingEnergyConsumption} developed two variants of the Restricted Boltzmann Machines (RBMs) stochastic model for forecasting residential energy consumption, namely the Conditional RBM (CRBM) and the Factored Conditional RBM (FCRBM). 
The authors compared the two methods with traditional machine learning methods such as Support Vector Machines (SVM), Artificial Neural Networks (ANN) and Recurrent Neural Networks (RNN), testing different time horizons with different time resolutions from one minute to one week. The FCRBM achieved the highest accuracy compared to ANN, RNN, SVM and CRBM. However, only one-dimensional time series were considered.


Recently, several hybrid deep learning models have been proposed. This concept consists of mixing different models with unique technique to address the limitations of a single model to increase the forecasting performance. For example, Convolutional Neural Network is used to extract spatial features and Recurrent Neural Networks is used to model temporal features.


Long Short-Term Memory Network (LSTM) models have been attracting the attention of many researches to perform energy consumption forecasting. Syed et al. \cite{Syed2021HSBUFC} proposed a framework consisting of a hybrid stacked bi-directional and uni-directional LSTM with fully connected dense layers and data cleaning process (HSBUFC). The prediction performance of this framework was superior than other hybrid deep learning models such as CNN-LSTM and ConvLSTM.

Kim and Chao \cite{Kim2019CNN-LSTM} integrated Convolutional Neural Network (CNN) with LSTM for forecasting the energy consumption of houses, namely CNN-LSTM. The CNN was used to extract features that affect energy consumption and the LSTM for modeling the temporal information. The hybrid model achieved better performance than LSTM and obtained values of 37.38 and 61.14 for MSE and RMSE, respectively. According to the authors, the main limitation of the hybrid model is the relatively large effort through trial and error to determine the optimal hyperparameters.


Sajjad et al. \cite{Sajjad2020CNN-GRU} implemented a hybrid deep learning-based energy forecasting model (CNN-GRU). Their solution uses CNN to extract spatial features from the energy consumption dataset and uses Gated Recurrent Unit (GRU) to exploit its effective gated structure to make energy consumption forecasting. The results showed that CNN-GRU had improved performance compared to individual forecasting models and the CNN-LSTM model.


Ullah et al. \cite{Ullah2020CNN-BI-LSTM} proposed a hybrid deep learning model by integrating CNN with multilayer bidirectional LSTM (M-BLSTM) for short-term power consumption forecasting and compared their results with bidirectional LSTM (BLSTM), LSTM and CNN-LSTM. The prediction result showed that the hybrid model outperforms LSTM, BLSTM and CNN-LSTM.


Munkhdalai et al. \cite{Munkhdalai2019AIS-RNN} presents the AIS-RNN model, which combines RNNs with an adaptive feature selection mechanism to improve forecasting performance. The model consists of two parts: the first model generates contextual importance weights for selecting appropriate features; then, the second model predicts the target variable (i.e., device energy consumption) based on the inputs. The authors compared their model with different machine learning and deep learning models such as LSTM, GRU, and SVM, and the results showed that the AIS-RNN model outperformed the other models.


Parhizkar et al. \cite{Parhizkar2021PCAEnergyConsuption} improved the performance of smart building forecasting models by using PCA to preprocess data and extract key features representing four smart building datasets for the subsequent forecasting task. The authors used five machine learning models (linear regression, support vector regression, regression tree, random forest, and K-nearest neighbours) to predict energy consumption.




\subsection{Fuzzy Time Series}
\label{fts}

The foundations of Fuzzy Time Series (FTS) were first proposed by Song and Chissom \cite{Song1993FTS} to deal with ambiguous and imprecise knowledge in time series data. FTS is a representation of time series using fuzzy sets as fundamental components, then conventional time series values are transformed into linguistic time series. Since the introduction of FTS, several categories of FTS methods have been proposed, which are distinguished by their order ($ \Omega $) and time variance. The order is the number of time delays (lags) used in modeling the time series. Time variance defines whether the FTS model changes over time \cite{Severiano2020NSFTS}. A first order FTS model requires $y(t-1)$ data to predict $y(t)$ and a high order FTS model requires $y(t - 1),...,y(t - p)$ data to predict $y(t)$.


In the training procedure of an FTS model, the Universe of Discourse ($U$) is partitioned into intervals bounded by the known limits of $Y$, where $U = [\min(Y), \max(Y)]$. For each interval, a fuzzy set $A_i \in \tilde{A} $ with its own membership function (MF) $\mu_{A_i}:\mathbb{R} \rightarrow [0,1]$ is defined, then each fuzzy set is assigned a linguistic value that represents a range of $U$. The crisp time series $Y$ is mapped to the fuzzified time series $F$, given the membership values to the fuzzy sets. Temporal patterns corresponding to the number of lags $ \Omega $ are created from $F$. Each pattern represents a fuzzy rule called Fuzzy Logical Relationship (FLR) and they are grouped by their same precedents to form a Fuzzy Logical Relationship Group (FLRG). The FLRG form the rule base which is the final representation of the FTS forecasting model.


Once the FTS model is trained, it can be used to predict new values. The crisp samples $y(t-\Omega), \ldots, y(t-1)$ are mapped to the fuzzified values $f(t-\Omega), \ldots, f(t-1)$, where $f(t) = \mu_{A_i}(y(t)), \forall A_i \in \tilde{A} $, for $t = 1, \ldots, T$. The rules matching the corresponding input are found. The FLRG whose precedent is equal to the input value is selected and the candidate fuzzy sets in its consequent are applied to estimate the predicted value.


\subsection{Dimensionality Reduction}
\label{dimensionality_reduction}

There are several approaches in the literature for dealing with high-dimensional data. Some of the main dimensionality reduction (embedding) techniques are feature selection and feature extraction. In feature selection, a subset of the original features is selected. On the other hand, in feature extraction, a set of new features is found by mapping from the existing input variables. The mapping can be either linear or non-linear.


The goal of embedding by feature extraction is to learn a function $\gamma:\mathbb{R}^{M} \rightarrow \mathbb{R}^{K}$ that maps $M$-dimensional features measured over $T$-time steps into the reduced $K$-dimensional feature space with $K~\ll~M$.


There are several feature extraction methods that can be used for different types of data and different requirements. In this paper, we focus on three of them: Principal Component Analysis (PCA) \cite{Pearson1901PCA}, Autoencoder (AE) \cite{Rumelhart1986Autoencoders} and Self-organizing maps (SOM) \cite{Kohonen2001SOM}.


\section{Embedding-based Fuzzy Time Series ($\gamma$FTS)}
\label{efts}

A multivariate time series $Y \in \mathbb{R}^M$ is an extension of the univariate case, containing values of different univariate time-dependent variables. The dimensionality of $Y$ is given by $M = |\mathcal{V}|$ and each vector (i.e., the set of variables) $y(t) \in Y$ carries all variables $\mathcal{V}_i \in \mathcal{V}$.


The multivariate FTS model is an extension of several univariate FTS models used to capture the linear or nonlinear interdependencies between multiple fuzzy time series. Multivariate FTS forecasting methods can be divided into Multiple Input Single Output (MISO) and Multiple Input Multiple Output (MIMO) methods. The MISO obtains a set of explanatory variables (or exogenous) $\mathcal{V}_i \in \mathcal{V}$ and only one target variable (or endogenous) $\mathcal{V}^* \in \mathcal{V}$, which is the output set, while the MIMO method uses the same input set of variables as the output set.


In this work, we propose a new MISO FTS method, called $\gamma$FTS, to handle high-dimensional multivariate time series by applying an embedding transformation, then reducing the dimensionality of the time series and enabling efficient pattern recognition and induction of fuzzy rules.


The $\gamma$FTS method is a data-driven and explainable method that is flexible and adaptable for many IoE applications. The proposed approach consists of embedding, training and forecasting procedures. This paper is a substantial extension of the works presented in \cite{Santos2021SOM-FTS} and \cite{Bitencourt2021ENSFTS}.


\subsection{Embedding Procedure}
\label{embedding}

The embedding procedure is responsible for extracting the principal components that better represent the content of the high-dimensional multivariate time series for the subsequent forecasting task.



\subsubsection{Principal Component Analysis (PCA)}
\label{pca}

Principal Component Analysis (PCA) is one of the most popular feature extraction approaches. PCA estimates the cross-correlation among the variables and extracts a reduced set of features, called principal components, which are linearly uncorrelated. The principal components explain the largest possible proportion of the data variance with the constraint that they are orthogonal to all previously extracted features. PCA extracts a reduced data representation that describes a specific total percentage of the data variance.


The PCA embedding procedure consists of the following steps: Starting from the multivariate time series $Y \in \mathbb{R}^{N \times M}$, we compute the covariance matrix $C \in \mathbb{R}^{M \times M}$ and extract the first $K$ eigenvectors related to the largest eigenvalues. Thus, we obtain the matrix $Z \in \mathbb{R}^{M \times K}$, which is used to compute the embedding feature $\gamma(x):Z^T \cdot y $ with $y \in \mathbb{R}^{M}$. 


\subsubsection{Autoencoder (AE)}
\label{AE}

Autoencoders (AEs) are a special type of artificial neural network whose task is to convert the input into a latent and meaningful representation, that is, to encode the data. The autoencoder was first introduced in \cite{Rumelhart1986Autoencoders} as a simple network that attempts to reconstruct the input, and it has been used to solve unsupervised learning and transfer learning problems, as this technique attempts to produce the best possible representation of the input by minimizing the reconstruction error. That is, it decodes back the encoded representation of the input and compares it with the real input \cite{Mohammadi2018SurveyDPIoTBigData} \cite{Gensler2016AE-LSTMPV}.


An autoencoder network consists of an encoder layer and a decoder layer connected by one or more hidden layers that transform high-dimensional data into a $K$-dimensional representation. The embedding transformation is achieved by exploiting bottleneck features extracted in the hidden layers, and the autoencoder can handle nonlinear correlations between variables.


The AE embedding procedure aims to learn the encoding function $e: \mathbb{R}^M \rightarrow \mathbb{R}^K$ and the decoding function $d: \mathbb{R}^K \rightarrow \mathbb{R}^M$, according to equation


\begin{equation}  \label{eq:ae_emb_1}
    (e(\cdot),d(\cdot)) = \arg \min_{e(\cdot),d(\cdot)} ||Y - d(e(Y))||^2 
\end{equation}

We have used the encoding function $e$ as a direct embedding function $\gamma$ for the high-dimensional time series $Y \in \mathbb{R}^{M}$, such that


\begin{equation}  \label{eq:ae_emb_2}
    \gamma(Y) = e(Y)
\end{equation}

\subsubsection{Self-organizing maps (SOM)}
\label{som}

Self-organizing maps (SOM) is an unsupervised learning approach that projects the original continuous space $\mathcal{D} \subseteq \mathbb{R}^M$ into a finite, discrete $K$-dimensional array W named grid, where each cell $w_i \in W$ corresponds to a point of $\mathcal{D}$, and each dimension $K$ has a fixed length $L \in \mathbb{N}^+$. The embedding space $\E$ of the map is composed of the $K$ coordinates of the grid, then its dimensions are finite, discrete and positive.


In the training procedure of a SOM model, the grid $W$ is gradually approximated to the topology of the original space by fitting the cells $w_i$ to the values of this space and their neighbor values. The training process is unsupervised and iterative. The SOM embedding procedure performs a nearest neighbor search on the grid $W$ to find the coordinates of the best matching cell. Formally, the embedding process transforms the input $y \in Y$ into the $k$-dimensional embedding space $y_{\gamma}(t)$


The training procedure is shown in Algorithm \ref{alg:som_training}, where $T$ is the length of the training data, $\alpha$ is the learning rate, $\sigma$ is the neighborhood radius, $d: \mathbb{R}^n \times \mathbb{R}^n \rightarrow \mathbb{R}^+ $ is the Euclidean distance metric, and $f(i, u, \sigma, \alpha)$ is the Gaussian neighborhood function given by \eqref{eq:gaussian_neigh}. The parameters $\alpha$ and $\sigma$ control the adaption velocity during training and the amplitude of the neighbor cells affected when $W$ is updated (using the Gaussian neighborhood function $f$), and both parameters are monotonically decreased during the training loop.


\begin{equation}
    f(i, u, \sigma, \alpha) = \alpha \cdot \exp\left( -\frac{||w_i - w_u||^2}{2\sigma^2} \right)    
\label{eq:gaussian_neigh}
\end{equation}

\begin{algorithm}[htb]
\small
\SetAlgoLined
\SetKwInOut{Input}{input}\SetKwInOut{Output}{output}
\Input{$Y$,  $k$,  $L$, $\alpha_0$, $\sigma_0$}
\Output{$W$}
Create the grid $W$ with $k$ dimensions and each dimension with $L$ cells, where each cell $w_i \in W$ is a $1\times n$ vector\;
Initialize each $w_i \in W$ with random values\;
\For{$t\leftarrow 1$ \KwTo $T$}{
    Get an instance $y(t) \in Y$\;
    Find the closest cell $w_u$ to $y(t)$ such that $w_u = \arg \min_{w_i}\; d(y(t), w_i)\; \forall  w_i \in W$\;
    \ForEach{$w_i \in W$}{
         $w_i \leftarrow w_i + f(i, u, \sigma_t, \alpha_t) \cdot (y(t) - w_i)$
    }
    Decrease the neighborhood radius $\sigma$ such that 
    $\sigma_t \leftarrow \sigma_0 \left( \frac{\sigma_t}{\sigma_0} \right)^\frac{t}{T}$ \\
    Decrease the learning rate $\alpha$ such that 
    $\alpha_t \leftarrow \alpha_0 \left( \frac{\alpha_t}{\alpha_0} \right)^\frac{t}{T}$
}
 \label{alg:som_training}
 \caption{SOM Training}
\end{algorithm}

The SOM training procedure is an iterative and unsupervised procedure that gradually approximates the grid $W$ to the topology of the original space by fitting the cells $w_i$ to the values of this space and their neighbor values.


The SOM embedding function is much simpler than the training procedure and simply performs a nearest neighbour search on the grid $W$ and returns the coordinates of the best matching cell. In this way, the map transforms the input $y \in Y$ into the $k$-dimensional embedding vector $y_\gamma(t)$. The SOM Embedding function is presented in Algorithm \ref{alg:som_embed}.


\begin{algorithm}[htb]
\small
\SetAlgoLined
\SetKwInOut{Input}{input}\SetKwInOut{Output}{output}
\Input{$y \in Y$,  $W$}
\Output{$y_{\gamma}$}
Find the closest cell $w_u$ to $y(t)$ such that \[w_u = \arg \min_{w_i}\; d(y(t), w_i)\; \forall  w_i \in W\]
$y_\gamma \leftarrow $ get map coordinates of $w_u$ in $W$\;
\label{alg:som_embed}
 \caption{SOM Embedding}
\end{algorithm}

\subsection{Training Procedure}
\label{training}

The training procedure shown in Figure~\ref{training_procedure} creates a multivariate FTS model $\mathcal{M}$ that captures all the information in the embedding data. Let the embedding multivariate time series $Y_{\gamma} \in \mathbb{R}^{K}$ and its individual instances $y_{\gamma}(t)$ $\in$ $Y_{\gamma}$ for $t = 0, 1,..., T$ and the number of fuzzy sets $\kappa$. The training procedure consists of the following steps:



\begin{figure}[htbp]
\begin{center}
 \includegraphics[width=8.5cm,height=4cm]{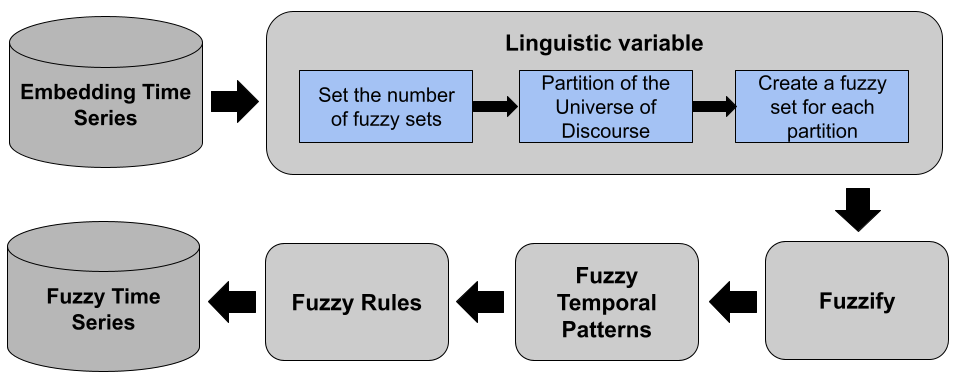}
  \caption{$\gamma$FTS Training procedure}
  \label{training_procedure}
\end{center}
\end{figure}

\begin{enumerate}
    \item \textbf{Partitioning}: define $U_{\mathcal{V_{\gamma}}}=[lb, ub]$, where $lb = \min(Y_{\gamma}) - D_1$ and $ub = \max(Y_{\gamma})+ D_2$, with $D_1 = r \times |\min(Y_{\gamma})|$ and $D_2 = r \times | \max(Y_{\gamma})|$, $0 < r < 1$. We extrapolate the known bounds of the variables $\mathcal{V}_{{\gamma}_i}$ as a security margin.
    
    \item \textbf{Defining the linguistic variable}: split $U_{\mathcal{V_{\gamma}}}$ in $\kappa_i$ (i.e. number of fuzzy sets) overlapping intervals $U_j$ with midpoints $c_j$ for $j = 0, 1, 2, ..., \kappa_i$. For each interval $U_j \in U_{\mathcal{V_{\gamma}}}$ create an overlapping fuzzy set $\mathcal{A}_{j}^{\mathcal{V}_i}$ with the membership function $\mu_{\mathcal{A}_j}^{\mathcal{V}_i}$ and generate a variable $\mathcal{V}_{{\gamma}_i}$ for the dimension $i$ and a linguistic variable 
 $\widetilde{\mathcal{V}}_{{\gamma}_i} \in \widetilde{\mathcal{V}}_{\gamma}$. Each fuzzy set $\mathcal{A}_{j}^{\mathcal{V}_i}$ represents a linguistic variable $\widetilde{\mathcal{V}}_{{\gamma}_i}$.
 
    \item \textbf{Fuzzification}: the embedding time series $Y_{\gamma}$ is then transformed into a fuzzy time series $F_{\gamma}$. Each data point $f_{\gamma}(t) \in F_{\gamma}$ is an $n \times \kappa$ array with the fuzzified values with respect to the linguistic variable, where the fuzzy membership is predefined as follows 
    
        \begin{equation} \label{eq:fuzzification}
            f(t)  \leftarrow  \left\{ \mathcal{A}_{j}^{\mathcal{V}_i} ~|~ \mu_{\mathcal{A}_j}^{\mathcal{V}_i} (Y_{\gamma}(t)^i) > 0 \right\} 
        \end{equation}
        
    \item \textbf{Generate the temporal patterns}: generate temporal patterns with the format $\mathcal{A}_{j}^{\mathcal{V}_0},...,\mathcal{A}_{j}^{\mathcal{V}_n} \rightarrow \mathcal{A}_{j}^{\mathcal{V}*}$, where the precedent or LHS (left hand side) is $f_{\gamma}(t) = \mathcal{A}_{j}^{\mathcal{V}_i}$ and the consequent or RHS (right hand side) is the target variable such that $f_{\gamma}(t+1) = \mathcal{A}_{j}^{\mathcal{V}^*}$. Both LHS and RHS are related to $\mathcal{A}_{j}^{\mathcal{V}_i}$ with maximum membership.
    
    \item \textbf{Generate the rule base}: finally, each pattern represents a fuzzy rule and they are grouped by their same precedents, creating a fuzzy rule $r$ with the format $LHS \rightarrow RHS$. Each fuzzy rule represents the set of possibilities with may happen on time $t + 1$ when a certain precedent is identified on previous lag. 
\end{enumerate}

\subsection{Forecasting Procedure}
\label{forecasting}

\begin{figure}[htbp]
\begin{center}
 \includegraphics[width=8.5cm,height=2cm]{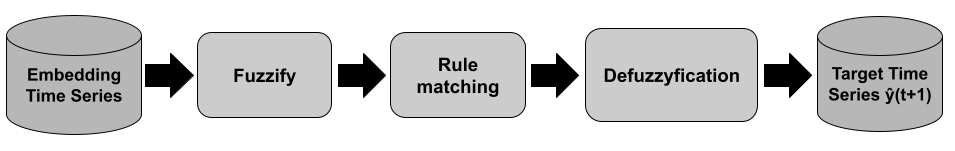}
  \caption{$\gamma$FTS Testing procedure}
  \label{testing_procedure}
\end{center}
\end{figure}

The forecasting procedure, shown in Figure~\ref{testing_procedure}, finds the rules that match a given fuzzified input and uses them to compute a numerical prediction using the fuzzy sets. This procedure aims to estimate $y(t+1)$ for the endogenous variable $\mathcal{V}^* \in \mathbb{R}$, given an input sample $y(t)_{{\gamma}_i}$ and using the fuzzy rules of the FTS model $\mathcal{M}$.



Given the embedding multivariate time series $Y_{\gamma} \in \mathbb{R}^{K}$ and its individual instances $y_{\gamma}(t)$ $\in$ $Y_{\gamma}$ for $t = 0, 1,..., T$, the following steps are taken to forecast $\hat{y}(t + 1)$. 

\begin{enumerate}
    \item \textbf{Fuzzification}: for each variable $\mathcal{V}_i \in \mathcal{V}$, we fuzzify the embedding data according to equation~\ref{eq:fuzzification}.
    
    \item \textbf{Rule matching}: select ${r}$ fuzzy rules whether any fuzzy set of $f(t)$ is equal to LHS. The rule fuzzy membership grade is computed using the minimum function T-norm as follows 

    \begin{equation} \label{eq:rule_matching}
        \mu_q =  \bigcap\limits_{j \in \widetilde{\mathcal{V}}_i ~;~ i \in \mathcal{V}} \mu_ji
    \end{equation}
    
    \item \textbf{Rule mean points}: for each rule $q$, we compute the mean point $mp_q$ of the endogenous variable $\mathcal{V}^*$ according the following equation 

    \begin{equation} \label{eq:mean_points}
        mp_q =  \sum\limits_{j \in \widetilde{\mathcal{V}}_{i}^*} c_j
    \end{equation}
    
    where $c_j$ is the $c$ parameter of the membership function from the fuzzy sets.
    
    \item \textbf{Defuzzification}: finally, the predicted value $\hat{y}(t + 1)$ is obtained as the weighted sum of the rule midpoints by their membership grades $\mu_{A_j}$, according to equation:

    \begin{equation} \label{eq:defuzzification}
         \hat{y}(t + 1) = \sum\limits_{q \in {r}} \mu_q \cdot mp_q
    \end{equation}
    
\end{enumerate}

\subsection{Embedding Weighted Multivariate Fuzzy Time Series ($\gamma$WMVFTS)}

To demonstrate our proposed method $\gamma$FTS, we extend the Weighted Multivariate Fuzzy Time Series (WMVFTS) method  \cite{e2019distributed} to enable it for high-dimensional time series (i.e. $\gamma$WMVFTS). The WMVFTS method is a weighted and rule-based MISO first-order multivariate method that allows individual partitioning schemes for each variable. The training procedure can be easily distributed in network clusters and its knowledge base is easy to understand and verify.


We used the embedding transformation presented in section~\ref{embedding} to reduce the dimensionality of the time series and enable efficient pattern recognition and induction of fuzzy rules. WMVFTS combined with PCA, AE and SOM are referred to as PCA-WMVFTS, AE-WMVFTS and SOM-WMVFTS, respectively.

\begin{figure*}[!h]
\begin{center}
 \includegraphics[width=17cm,height=7cm]{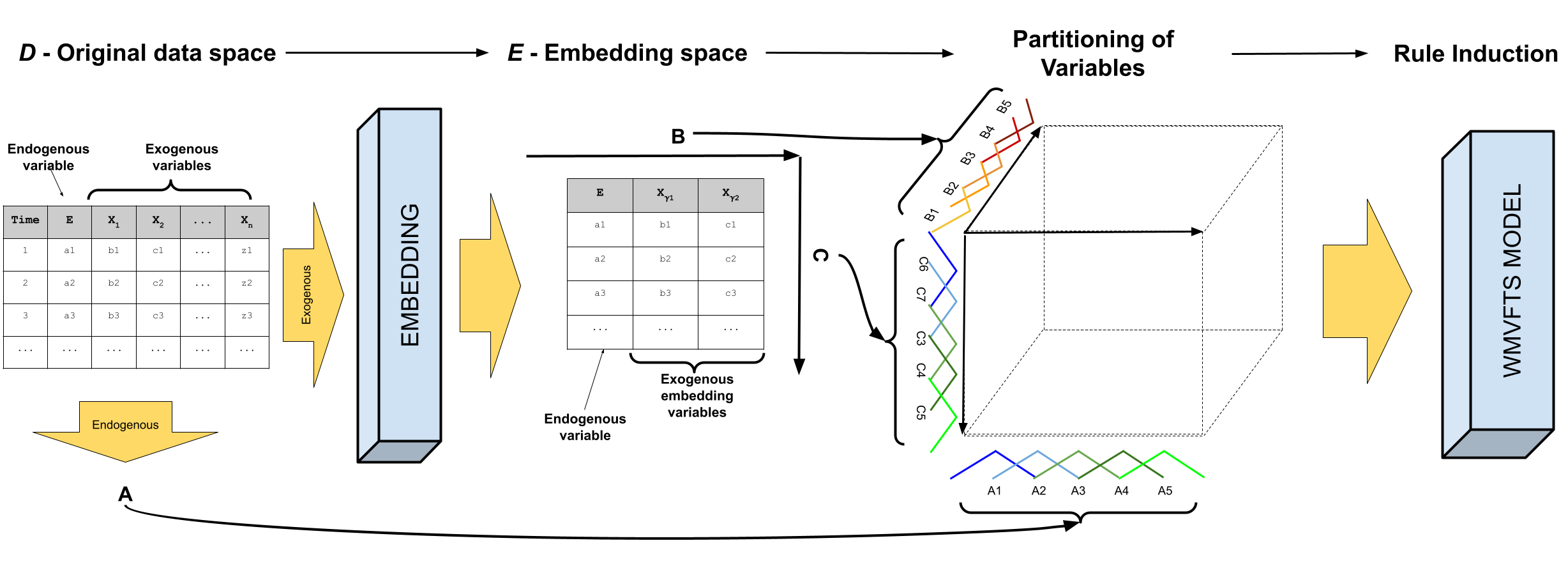}
  \caption{$\gamma$WMVFTS training for number of dimensions $k = 2$ and number of fuzzy sets $\kappa = 5$}
  \label{model_training}
\end{center}
\end{figure*}

The training  procedure of $\gamma$WMVFTS model is presented in Algorithm~\ref{alg:ewmvfts_training} and illustrated in Figure~\ref{model_training}. In the \textbf{Generate the rule base} step, we create fuzzy rules $r$ with the format $LHS \rightarrow w   \cdot  RHS$, where $w = |\mathcal{A}_{j}^{\mathcal{V}_i}| / |RHS|$ which are normalized frequencies of each temporal pattern (i.e. weights) according to equation 

\begin{equation} \label{eq:fuzzy_rules_weights}
  w_i = \frac{\#\mathcal{A}_{j}^{\mathcal{V}^*}}{\#RHS} ~~~~ \forall \mathcal{A}_{j}^{\mathcal{V}^*} \in RHS
\end{equation}
where $\#\mathcal{A}$ is the number of occurrences of $\mathcal{A}_i$ on temporal patterns with the same LHS and $\#RHS$ is the total number of temporal patterns with the same precedent LHS. The other steps remain the same as described in Section~\ref{training}.

\begin{algorithm}[htb]
\small
\SetAlgoLined
\SetKwInOut{Input}{input}\SetKwInOut{Output}{output}
\Input{A multivariate time series $Y$, number of instances $N$, number of dimensions $K$, number of fuzzy sets $\kappa$}
\Output{the linguistic variable $\widetilde{\mathcal{V}}$, the rule set $\mathcal{R}$}
\ForEach{$y(t) \in Y$}{
	$y_{\gamma}(t)\leftarrow$ Embedding Procedure($y(t)$, $K$)\;
	$Y_{\gamma} \leftarrow y_{\gamma}(t)$
} 
\For{$i \leftarrow 1 \ldots k$}{
    $u  \leftarrow$ Split $UoD$ in $\kappa$ overlapping intervals\;
    $\mathcal{V}_i \leftarrow$ Create a variable for the dimension $i$\;
    $\widetilde{\mathcal{V}} \leftarrow$ Create an empty linguistic variable for $\mathcal{V}_i$\;
    \For{$j \leftarrow 1 \ldots N_i$}{
        $\mathcal{A}_{j}^{\mathcal{V}_i} \leftarrow$ Create a fuzzy set with the interval $u_j$ with membership function $\mu_{\vari}$\;
        $\widetilde{\mathcal{V}} \leftarrow \mathcal{A}_{j}^{\mathcal{V}_i} $\;
    }
}
$F \leftarrow $ Create empty fuzzified data\;
\ForEach{$y_\E(t) \in Y_\E$}{
	$f(t) \leftarrow  \{ \fset \; |\; \mu_{\fset}(y_\E(t)^{i}) > 0\}$ for all fuzzy sets $\fset \in \lvari$ and all variables $\vari$\;
	$F \leftarrow f(t)$\;
}
$\mathcal{R} \leftarrow $ Create an empty set of rules \;
\ForEach{$f(t), f(t+1) \in F$}{
    \ForEach{ fuzzy set $LHS \in f(t)$}{
        $RHS \leftarrow$ all fuzzy sets $\fset$ in $f(t+1)$ with weight $w = |\fset|/|RHS|$\;
        $r \leftarrow$ Create rule $LHS \rightarrow RHS$\;
        \If{$r$ does not exist in $\mathcal{R}$}{
            $\mathcal{R} \leftarrow r$; 
        }
    }
}
\label{alg:ewmvfts_training}
\caption{$\gamma$WMVFTS Training}
\end{algorithm}

Algorithm~\ref{alg:ewmvfts_forecasting} shows the forecasting  procedure of the $\gamma$WMVFTS model. In \textbf{Rule mean points} step, for each rule $q$, we compute the mean point $mp_q$ of the endogenous variable $\mathcal{V}^*$ as follows  

\begin{equation} \label{eq:mean_points}
    mp_q =  \sum\limits_{j \in \widetilde{\mathcal{V}}_{i}^*} w_j \cdot c_j
\end{equation}
where $w_j$ is the weights calculated according to~\ref{eq:fuzzy_rules_weights}. The other steps remain the same as presented in Section~\ref{forecasting}. 

\begin{algorithm}[htb]
\small
\SetAlgoLined
\SetKwInOut{Input}{input}\SetKwInOut{Output}{output}
\Input{A sample $y(t)$, the dimensions $K$, the linguistic variable $\widetilde{\mathcal{V}}$, the rule set $\mathcal{R}$}
\Output{target forecasting value $\hat{y}(t+1)$}
$y_{\gamma}(t)\leftarrow$ Embedding Procedure($y(t)$, $K$)\;
$f(t) \leftarrow  \{ \fset \; |\; \mu_{\fset}(y_\E(t)^{i}) > 0\}$ for all fuzzy sets $\fset \in \lvari$ and all variables $\vari$\;
$\mathcal{R}_{fired} \leftarrow $ Create an empty set of rules\;
$\mu \leftarrow $ Create the empty vector of activation of each rule\;
\ForEach{$r \in \mathcal{R}$}{
    \If{any fuzzy set of $f(t)$ is equal to $r_{LHS}$}{
        $\mathcal{R}_{fired} \leftarrow r$\;
    }
}
\ForEach{$r \in \mathcal{R}_{fired}$}{
    $mp_r \leftarrow \sum_{\fset \in r_{RHS}} w_i \cdot mp_i$ where $w_i$ is the weight and $mp_i$ is the midpoint of the fuzzy set\;
    $\mu_r \leftarrow \bigcap_{\fset \in r_{LHS}} \mu_{\fset}(y(t))$
}
$\hat{y}(t+1) = \frac{\sum_{r \in \mathcal{R}_{fired}} \mu_r \cdot mp_r}{\sum_{r \in \mathcal{R}_{fired}} \mu_r}$
\label{alg:ewmvfts_forecasting}
\caption{$\gamma$WMVFTS Forecasting}
\end{algorithm}

\section{Experiments}
\label{experiments}

Several experiments have been conducted to obtain highly accurate energy forecasting in smart buildings using the proposed methodology. We compare the performance of our proposed models with the baseline models and state-of-the-art forecasting models proposed in the literature.

This section includes the description of the case of studies, the methodology of the experiments, the evaluation metrics, the description of the baseline model, the optimization of the hyperparameters, the computational experiments and the reproducibility.



\subsection{Case studies}
\label{case_studies}

An important application of IoE in smart buildings is monitoring the energy consumption of devices. This importance stems from the fact that proper monitoring of energy appliances can reduce power consumption and provide better energy and cost savings. In addition, energy forecasting at the customer level will reflect directly into efficiency improvements across the power grid.

As an example of the approach presented here, we evaluate the proposed methodology on three datasets presented below. The input data of each dataset was refined before training our $\gamma$WMVFTS models. Therefore, we employ data pre-processing strategies to remove outliers and missing values on datasets. 



\subsubsection{UCI Appliances Energy Consumption (AEC-DS)}
\label{uci_appliances}

The ``UCI Appliances energy prediction" dataset \cite{Dua2019UCIDatasets} includes measurements of temperature and humidity collected from a WSN, weather information from a nearby Weather Station, and recorded energy consumption from appliances and lighting fixtures. The energy appliances data were obtained by taking continuous measurements (every 10 minutes) in a low-energy house in Belgium over a period of 137 days between January 11, 2016 and May 27, 2016 (approximately 4.5 months). The dataset contains 19,735 instances, including 26 explanatory variables and 1 temporal variable (date/time). Table~\ref{tab:uci_appliances} shows all variables.

\begin{table}[htb]
\centering\small
\caption{UCI Appliances Energy Consumption Dataset (AEC-DS)}
\begin{tabular}{cc} 
\toprule
\textbf{Variables} & \textbf{Description }             \\ 
\midrule
Appliances         & Appliances energy consumption     \\ 

lights             & Light energy consumption          \\ 

RH\_1              & Humidity in kitchen area          \\ 

T2                 & Temperature in living room area   \\ 

RH\_2              & Humidity in living room area      \\ 

T3                 & Temperature in laundry room area  \\ 

RH\_3              & Humidity in laundry room area     \\ 

T4                 & Temperature in office room        \\ 

RH\_4              & Humidity in office room           \\ 

T5                 & Temperature in bathroom           \\ 

RH\_5              & Humidity in bathroom              \\ 

T6                 & Temperature outside the building  \\ 

RH\_6              & Humidity outside the building     \\ 

T7                 & Temperature in ironing room       \\ 

RH\_7              & Humidity in ironing room          \\ 

T8                 & Temperature in teenager room 2    \\ 

RH\_8              & Humidity in teenager room 2       \\ 

T9                 & Temperature in parents room       \\ 

RH\_9              & Humidity in parents room          \\ 

T\_out             & Temperature outside               \\ 

Press\_mm\_hg      & Pressure                          \\ 

RH\_out            & Humidity outside                  \\ 

Windspeed          & Wind Speed                        \\ 

Visibility         & Visibility                        \\ 

Tdewpoint          & Dew Point                         \\ 

date               & Date time stamp                   \\
\bottomrule
\end{tabular}
\label{tab:uci_appliances}
\end{table}

The appliances energy consumption (Wh) measured is the focus of our analysis, then it was chosen as the target variable (endogenous variable) $\mathcal{V}^*$ and the set of explanatory variables (exogenous variable) $\mathcal{V}$ is composed of 24 variables.

\subsubsection{UCI Household Power Consumption (HPC-DS)}
\label{uci_household}

The ``UCI Individual household electric power consumption" dataset \cite{Dua2019UCIDatasets} contains electricity consumption data from a residential house in France. Electricity consumption was collected by continuous measurements over a four-year period from December 2006 to November 2010 with a resolution of one minute. In this work, in some experiments we changed the time resolution of this dataset, growing it to 30 minutes. The dataset consists of 2,075,259 instances and 12 variables, including 7 explanatory variables and 2 temporal variables. A total of 25,979 missing values were removed in pre-processing. Submetering indicate electricity consumption in the kitchen, laundry room, and for the air conditioner and electric water heater. Table~\ref{tab:uci_household} shows all variables.


The focus of our analysis is the measured global active power (kW), which we have chosen as the endogenous variable $\mathcal{V}^*$ and the set of exogenous variables $\mathcal{V}$ is composed of the past values of six variables.


\begin{table}[htb]
\centering\small
\caption{UCI Household Power Consumption Dataset (HPC-DS)}
\begin{tabular}{cc}
\toprule
\textbf{Variables}      & \textbf{Description }         \\ 
\midrule
date                    & Date                          \\ 

time                    & time                          \\ 

global\_active\_power   & global active power           \\ 

global\_reactive\_power & global reactive power         \\ 

voltage                 & averaged voltage              \\ 

global\_intensity       & ~global current intensity     \\ 

sub\_metering\_1        & energy sub-metering No. 1.    \\ 

sub\_metering\_2        & energy sub-metering No. 2.~   \\ 

sub\_metering\_3        & energy sub-metering No. 3. .  \\
\bottomrule
\end{tabular}
\label{tab:uci_household}
\end{table}

\subsubsection{Kaggle Smart Home with Weather Information Data\-set (SHWI-DS)}
\label{kaggle_smart_home}

The ``Kaggle Smart home with Weather Information" \cite{KaggleDataset} is a public dataset of IoE devices for smart buildings with weather capabilities. The dataset contains home appliance consumption in kW and weather data from January 2016 to December 2016 with a frequency of 1 minute (in this work, we changed the time resolution to 10 minutes). The dataset contains 500,910 instances and 29 variables, including 18 electricity data features, 10 weather data features, and 1 temporal data feature. Table~\ref{tab:kaggle_smart_home} shows all variables.


The total measured energy consumption (kW) is our target variable $\mathcal{V}^*$ and the set of exogenous variables $\mathcal{V}$ is composed of the remaining variables.


\begin{table}[htb]
\centering\small
\caption{Kaggle Smart Home with Weather Information Dataset (SHWI-DS)}
\begin{tabular}{cp{5.0cm}} 
\toprule
\textbf{Variables}  & \textbf{Description }              \\ 
\midrule
use                 & Total energy consumption           \\ 

gen                 & Total energy generated by solar~   \\ 

Dishwasher          & Energy consumed by the dishwasher  \\ 

Furnace 1           & Energy consumed by furnace 1       \\ 

Furnace 2           & Energy consumed by furnace 2       \\ 

Home office         & Energy consumed in home office     \\ 

Fridge              & Energy consumed by fridge          \\ 

Wine cellar         & Energy consumed by wine cellar     \\ 

Garage door         & Energy consumed by garage door     \\ 

Kitchen 12          & Energy consumed in kitchen 1       \\ 

Kitchen 14          & Energy consumed in kitchen 2       \\ 

Kitchen 38          & Energy consumed in kitchen 3       \\ 

Barn                & Energy consumed by barn            \\ 

Well                & Energy consumed by well            \\ 

Microwave           & Energy consumed by microwave       \\ 

Living room         & Energy consumed in living room     \\ 

Solar               & Solar power generation             \\ 

temperature         & Temperature                        \\ 

humidity            & Humidity                           \\ 

visibility          & Visibility                         \\ 

apparentTemperature & Apparent Temperature               \\ 

pressure            & Pressure                           \\ 

windSpeed           & Wind Speed                         \\ 

windBearing         & Wind Direction                     \\ 

dewPoint            & Dew Point                          \\ 

precipProbability   & Precipitation Probability          \\ 

time                & Date time stamp                    \\
\bottomrule
\end{tabular}
\label{tab:kaggle_smart_home}
\end{table}

\subsection{Experiments Methodology}
\label{experiments_methodology}

\begin{figure}[htbp]
\begin{center}
 \includegraphics[width=8.5cm,height=4cm]{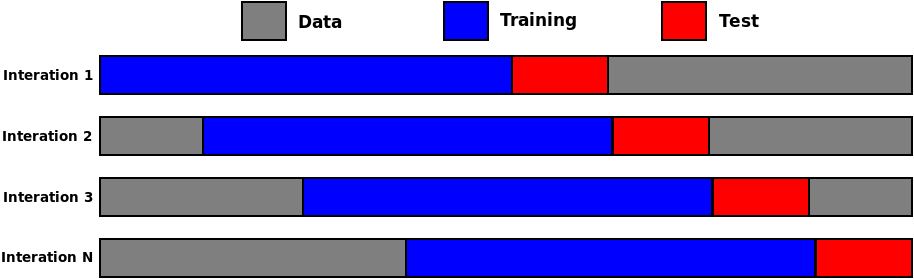}
  \caption{Schematic of the sliding window cross-validation.}
  \label{sliding_window}
\end{center}
\end{figure}

We separate 75\% of the data for the training set and 25\% for testing and use sliding window cross-validation in the computational experiments. The sliding window is a re-sampling technique based on splitting the data set into multiple training and testing subsets. The overall forecasting accuracy is obtained by considering the metric measures over all test subsets. Figure~\ref{sliding_window} shows cross-validation with sliding window.


The number of instances $N$ of each dataset was divided into 30 data windows with $N_{test}$ instances. For each window, we train the forecasting models using the training set, apply the model to the test set, and compute the prediction metrics over the test set. So for each model, there are 30 experiments and we evaluate the performance of the proposed method using the average error value measured in all the windows used for forecasting in the experiments. The performance metrics are described below.


\subsubsection{Performance metrics}
\label{perfomance_metrics}

In order to assess the accuracy of each model, the statistical metrics below were analyzed. Let $\hat{y}_i$ be the predicted values at the point of interest at time $t$, $y_{i}$ the observed values and $N$ the length of the dataset.

\begin{itemize}
    \item Root Mean Squared Error (RMSE), defined by equation \ref{eq:rmse}. It shows how accurate the forecasting model is, as it compares the difference between the predicted value and the real value (error), returning the standard deviation of this difference.
    
    \begin{equation}\label{eq:rmse}
        RMSE = \sqrt{\frac{\sum_{i=1}^{N}{(y_i - \hat{y}_i)^2}}{N}}
    \end{equation}
    
    \item Mean Absolute Percentage Error (MAPE), defined by equation \ref{eq:mape}. Shows accuracy in terms of percentage error.

    \begin{equation}\label{eq:mape}
        MAPE = \frac{1}{N}\sum_{i=1}^{N}{\left|\frac{(y_i - \hat{y}_i)}{y_i}\right|}
    \end{equation}
    
    \item Mean Absolute Error (MAE), defined by equation. It demonstrates the percentage difference between predicted values.
   
   \begin{equation}
        MAE = \frac{1}{N}\sum_{i=1}^{N}\left | y_{i} - \hat{y}_i \right |
   \end{equation} 
    
    \item Symmetric Mean Absolute Percentage Error (SMAPE), defined by equation 

    \begin{equation}\label{eq:smape}
        SMAPE = \frac{1}{N}\sum_{i=1}^{N}{\left|\frac{(y_i - \hat{y}_i)}{|\hat{y}_i| + |y_i|}\right|}
    \end{equation}
    
\end{itemize} 

In addition to the performance evaluation metrics presented above, we evaluate the performance of our proposed methodology using the Skill Score Index. The skill score defines the difference between the forecast and the reference forecast:
\begin{equation} \label{eq:skillScore}
Skill Score = 1 - \frac{M_{F}}{M_{R}}
\end{equation}
where $M_F$ refers to the value of the metric for the forecasting method and $M_R$ is the value of the metric for the reference method. The Skill Score can be used not only to compare with a naive model, but also to compare different forecasting methods with each other \cite{Cyril2017MLPV}. For example, a skill score of 0.50 indicates a 50\% improvement in accuracy over the competing model. A negative score indicates worse performance than the competitor model.

\subsection{Hyperparameters}
\label{hyperparameters_optimization}


This section is about finding the right combination of hyperparameters values for Autoencoder and SOM that help us find either the minimum error or the maximum accuracy.


To find the best configuration for embedding with Autoencoder, we empirically tested seven different hyperparameters: Epochs, Stack Size, Optimizer, Learning Rate, Kernel Initializer, Number of Neurons, and Activation Function of Neurons. Each of these hyperparameters was evaluated in a two-layer AutoEncoder architecture with a $l1$ regularizer in the first layer of both the encoder and the decoder. 

In relation to SOM, we also empirically tested the learning rate, epochs, and grid size (i.e. the maximum value of each data in the projected dataset). In addiction, the initial neighborhood radius used for all experiments was 12.5. 


Since the initialization of the weights is random, the results may be slightly different each time the program is run for both algorithms. Therefore, the mean RMSE and MAPE values of 10 forecast iterations were registered for each hyperparameter to compute a better approximation of each value. 

Table~\ref{tab:ae_hp} shows the values tested for each hyperparameter with autoencoder and the values that obtained the best result are shown in boldface. Table \ref{tab:som_hp} show tested and final (boldface) hyperparameters for SOM.

\begin{table}[htb]
\caption{Hyperarameters tested for the AutoEncoder. The best configuration is shown in boldface. }
\centering\small
\begin{tabular}{cp{5.0cm}}
\toprule
\textbf{Hyperparameter} & \textbf{Values}         \\ 
\midrule
Epochs             & 10, \textbf{50}, 100     \\
Batch Size         & 10, 20, \textbf{40}, 60, 80, 100     \\
Optimizer           & Adam, SGD, RMSprop, AdaDelta, \textbf{AdaGrad}, AdaMax, NAdam    \\
Learning Rate      & 0.001, 0.01, \textbf{0.1}, 0.2, 0.3       \\
Kernel Initializer & leCun Uniform, \textbf{Normal},  Glorot Normal, Glorot Uniform, He Normal, He Uniform   \\
Neurons            & 10, 15, 20, 30, 40, \textbf{50}, 100        \\
Activation         & Linear, Softmax, Softplus, \textbf{Softsign}, Tanh, Sigmoid, Hard Sigmoid              \\
\bottomrule
\label{tab:ae_hp}
\end{tabular}
\end{table}

\begin{table}[htb]
\caption{Hyperarameters tested for the SOM. The best configuration found is shown in boldface.}
\centering\small
\begin{tabular}{cp{5.0cm}}
\toprule
\textbf{Hyperparameter} & \textbf{Values}         \\ 
\midrule
Epochs             & \textbf{70}, 80, 90, 100     \\
Learning Rate      & \textbf{0.00001}, 0.001, 0.01, 0.1,        \\
Grid Size          & \textbf{20}, 30, 50, 100        \\
\bottomrule
\label{tab:som_hp}
\end{tabular}
\end{table}



\subsection{Baseline models}
\label{baseline_models}

We compared the performance of our proposed approach with the following baseline models: SARIMAX, PCA-SARIMAX, LSTM and persistence (naive) model, which is a reference technique that assumes that $y(t)$ equals $y(t-1)$.


The SARIMAX$(p,d,q)(P,D,Q)$ model (Seasonal Auto Regressive Integrated Moving Average with eXogenous variables) is an extension of SARIMA (seasonal ARIMA) with the possibility of integrating explanatory variables, where $p,d,q$ and $P,D,Q$ are non-negative integers related to the polynomial order of the autoregressive (AR), integrated (I), and moving average (MA) non-seasonal and seasonal components, respectively.


The selection of seasonal $(P,D,Q)$ and non-seasonal $(p,d,q)$ components was based on the Akaike Information Criterion (AIC) using the \texttt{auto\_arima} function from the pmdarima library \cite{pmdarima}. \texttt{auto\_arima} performs a grid search over multiples values of seasonal $(P,D,Q)$ and non-seasonal $(p,d,q)$ and returns the model with the lower AIC value. We also analysed the autocorrelation (ACF) and partial autocorrelation (PACF) plots to define the seasonal and non-seasonal components. Table~\ref{sarimax_componets} shows the values used in each dataset and the input data was normalized before model training, but the forecast error was computed based on actual energy consumption value. 


\begin{table}[htb]
\centering
\caption{Seasonal and non-seasonal components used on SARIMAX model}
\begin{tabular}{cc}
\toprule
\textbf{Dataset} & \textbf{SARIMAX(p,d,q)(P,D,Q,m)}  \\
\midrule
AEC-DS           & (0,0,1)(0,1,1,7)                  \\
HPC-DS           & (1,0,1)(1,1,1,7)                 \\
SHWI-DS          & (0,0,1)(0,1,1,7)                  \\
\bottomrule
\label{sarimax_componets}
\end{tabular}
\end{table}



Using the PCA algorithm presented in Section~\ref{pca}, we transform $M$ features of each dataset into $K = 2$ features and apply the SARIMAX model (henceforth called PCA-SARIMAX) described above. Seasonal and non-seasonal components were selected based on the $K$ dimensions, and Table~\ref{pca_sarimax_componets} shows the values used in each dataset.

\begin{table}[htb]
\centering
\caption{Seasonal and non-seasonal components used on PCA-SARIMAX model}
\begin{tabular}{cc}
\toprule
\textbf{Dataset} & \textbf{SARIMAX(p,d,q)(P,D,Q,m)}  \\
\midrule
AEC-DS           & (1,0,0)(1,0,0,7)                  \\
HPC-DS           & (1,0,1)(1,1,1,7)                 \\
SHWI-DS          & (1,0,0)(2,0,1,7)                  \\
\bottomrule
\label{pca_sarimax_componets}
\end{tabular}
\end{table}


A Long Short-Term Memory \cite{Hochreiter1997LSTM} is a recurrent neural network capable of learning to represent large datasets, which makes it extremely useful for applications such as time series forecasting. The LSTM architecture used consists of three layers and kernel regularizers, and we used the configuration of hyperparameters shown in Table~\ref{lstm_hp} for all datasets. These hyperparameters were chosen using hyperopt \cite{Bergstra2013Hyperopt}, which is a library designed for hyperparameter tuning. In other words, it searches for the hyperparameters that provide the minimum loss. Several combinations were tested and we used the best one as baseline.


Since the neural networks are sensitive to diverse data, we normalized the input data before training the LSTM model, but the forecast error was calculated based on the actual energy consumption.

\begin{table}[htb]
\centering
\caption{LSTM Hyperparameters}
\begin{tabular}{cc}
\toprule
\textbf{Hyperparameter} & \textbf{Value}  \\
\midrule
Neurons           & [50,30,10]                  \\
Activation         & ReLU                  \\
Optimizer         & ADAM                  \\
Batch Size       & 10                   \\
Epochs        & 70                    \\
Loss Function  & MSE \\
\bottomrule
\label{lstm_hp}
\end{tabular}
\end{table}

\subsection{Computational Experiments and Reproducibility}
\label{reproducibility}

All proposed $\gamma$WMVFTS models were implemented and tested using the programming language Python 3, and the open-source pyFTS \cite{pyFTS} and Scikit-Learn \cite{scikit-learn} libraries. The baseline models were implemented using Python 3, and the following open-sources libraries: Keras \cite{keras}, Tensorflow \cite{tensorflow} and StatsModels \cite{statsmodels}.  

To promote the transparency and the reproducibility of results, all proposed models are available at open-source pyFTS.


\section{Results}
\label{results}


This section presents the experimental results of our proposed models over several forecasting models tested on the AEC-DS, HPC-DS, and SHWI-DS datasets. First, we present the forecast performance of PCA-WMVFTS and AE-WMVFTS over different configurations of $K$ dimensions and the partitioning of the target variable (i.e. number of fuzzy sets), comparing a linear embedding algorithm with a nonlinear embedding algorithm. Second, we compare the forecast results of our models with the baseline models described in Section~\ref{baseline_models}. Third, we compare the performance of our proposed models with several state-of-the-art forecasting models proposed in the literature. Finally, we present a discussion about the parsimony, computational cost and explainability of our proposed methodology.

\begin{table*}[!htb]
\centering
\caption{PCA-WMVFTS and AE-WMVFTS model performance with different number of $K$ dimensions (DIM) and number of fuzzy sets (FS) on AEC-DS}
\begin{tabular}{ccp{1.6cm}p{1.6cm}p{1.6cm}p{2.0cm}p{1.6cm}p{1.6cm}p{1.5cm}p{1.5cm}}
\toprule
    &    & \multicolumn{4}{c}{\textbf{PCA-WMVFTS}}   & \multicolumn{4}{c}{\textbf{AE-WMVFTS}}     \\
\midrule
DIM & FS & RMSE(Wh) & MAE(Wh) & MAPE(\%) & SMAPE(\%) & RMSE(Wh) & MAE(Wh) & MAPE(\%) & SMAPE(\%)  \\
\midrule
2   & 10 & 28.957    & 20.513   & 29.81     & 11.582     & 31.388    & 23.469   & 36.062    & 13.352      \\
   & 20 & 17.095    & 8.683    & 11.233    & 4.694      & 18.793    & 10.325   & 14.334    & 5.89        \\
   & 30 & 9.859     & 4.326    & 5.212     & 2.225      & 12.295    & 5.714    & 6.965     & 2.934       \\
   & 40 & 8.585     & 2.897    & 3.163     & 1.285      & 10.802    & 4.315    & 5.063     & 2.113       \\
   & 50 & \textbf{5.457}     & \textbf{2.05}     & \textbf{1.758}     & \textbf{0.746}      & \textbf{5.516}     & \textbf{2.493}    & \textbf{2.405}     & \textbf{1.107}       \\
\hline
3   & 10 & 16.988    & 9.556    & 14.354    & 5.401      & 20.251    & 12.914   & 19.361    & 7.428       \\
   & 20 & 4.117     & 2.158    & 2.166     & 0.973      & 9.689     & 4.09     & 4.433     & 1.821       \\
   & 30 & 2.06      & 1.316    & 0.671     & 0.298      & 2.948     & 1.636    & 1.005     & 0.448       \\
   & 40 & 0.996     & 1.072    & 0.286     & 0.131      & 1.609     & 1.239    & 0.601     & 0.28        \\
   & 50 & \textbf{0.348}     & \textbf{0.962}    & \textbf{0.107}     & \textbf{0.049}      & \textbf{0.781}     & \textbf{1.029}    & \textbf{0.216}     & \textbf{0.103}       \\
\hline
4   & 10 & 10.451    & 4.4      & 6.11      & 2.069      & 11.872    & 5.782    & 7.356     & 2.968       \\
   & 20 & 1.495     & 1.182    & 0.507     & 0.215      & 1.663     & 1.304    & 0.549     & 0.255       \\
   & 30 & 0.415     & 0.958    & 0.093     & 0.041      & 0.792     & 1.033    & 0.201     & 0.086       \\
   & 40 & 0.187     & 0.935    & 0.061     & 0.026      & 0.137     & 0.925    & 0.044     & 0.019       \\
   & 50 & \textbf{0.083}     & \textbf{0.92}     & \textbf{0.035}     & \textbf{0.015}      & \textbf{0.078}     & \textbf{0.917}    & \textbf{0.033}     & \textbf{0.013}       \\
\hline
5   & 10 & 6.006     & 2.561    & 3.086     & 0.872      & 5.314     & 3.003    & 2.921     & 1.209       \\
   & 20 & 0.587     & 0.991    & 0.157     & 0.061      & 0.407     & 0.988    & 0.124     & 0.059       \\
   & 30 & 0.071     & 0.917    & 0.028     & 0.011      & 0.099     & 0.922    & 0.039     & 0.017       \\
   & 40 & 0.042     & 0.913    & 0.024     & 0.009      & 0.047     & 0.913    & 0.025     & 0.01        \\
   & 50 & \textbf{0.044}     & \textbf{0.913}    & \textbf{0.024}     & \textbf{0.009}      & \textbf{0.069}     & \textbf{0.915}    & \textbf{0.026}     & \textbf{0.01}        \\
\hline
6   & 10 & 2.697     & 1.566    & 1.301     & 0.452      & 2.858     & 1.549    & 0.902     & 0.418       \\
   & 20 & 0.11      & 0.923    & 0.037     & 0.016      & 0.145     & 0.93     & 0.047     & 0.02        \\
   & 30 & 0.051     & 0.914    & 0.025     & 0.01       & 0.067     & 0.916    & 0.029     & 0.012       \\
   & 40 & 0.041     & 0.913    & 0.024     & 0.009      & 0.041     & 0.913    & 0.024     & 0.009       \\
   & 50 & \textbf{0.039}     & \textbf{0.912}    & \textbf{0.024}     & \textbf{0.009}      & \textbf{0.039}     & \textbf{0.912}    & \textbf{0.024}     & \textbf{0.009}        \\     
\bottomrule
\label{tab:dim_fs_aec}
\end{tabular}
\end{table*}

\subsection{Embedding dimensions and number of fuzzy sets}
\label{dimensions_fuzzy_sets}

Tables~\ref{tab:dim_fs_aec}, \ref{tab:dim_fs_shwi} and \ref{tab:dim_fs_hpc} present the model performance of PCA-WMVFTS and AE-WMVFTS models over a varying number of $K$ dimensions and $\kappa$ number of fuzzy sets on datasets AEC-DS, SHWI-DS, and HPC-DS, respectively. The main hyperparameters of the proposed models are the embedding dimension $K$ and the number of fuzzy sets $\kappa$, then we can improve the performance of our models by just increasing the values of these hyperparameters. The performance metrics were computed in the testing set.

Regarding to AEC-DS, PCA-WMVFTS is just slightly superior than AE-WMVFTS in the most of combinations of $K$ and $\kappa$ for all the accuracy metrics, but not significantly. Both embedding methods achieved similar performance. Therefore, our forecasting methods are equally good in most scenarios. However, the time spent to train the autoencoder has to be taken into account, since it is a neural network and it takes a considerable time to train and optimize the reconstruction error.

In contrast, in relation to SHWI-DS dataset, PCA-WMVFTS showed a superior performance than AE\--WMV\-FTS in all the combinations of number of fuzzy sets and reduced dimensions. As the number of $K$ dimensions and $\kappa$ fuzzy sets increase, the difference between the methods decrease. Besides, our models presented optimal prediction results with a small number of reduced dimensions. 

\begin{table*}[!h]
\centering
\caption{PCA-WMVFTS and AE-WMVFTS model performance with different number of K-dimensions (DIM) and fuzzy sets (FS) on SHWI-DS}
\begin{tabular}{ccp{1.6cm}p{1.6cm}p{1.6cm}p{2.0cm}p{1.6cm}p{1.6cm}p{1.5cm}p{1.5cm}}
\toprule
    &    & \multicolumn{4}{c}{\textbf{PCA-WMVFTS}}   & \multicolumn{4}{c}{\textbf{AE-WMVFTS}}     \\
\toprule
DIM & FS & RMSE(Wh) & MAE(Wh) & MAPE(\%) & SMAPE(\%) & RMSE(Wh) & MAE(Wh) & MAPE(\%) & SMAPE(\%)  \\
\midrule
2   & 10 & 0.424     & 0.306    & 177.341   & 18.895     & 0.496     & 0.367    & 238.393   & 21.833      \\
   & 20 & 0.31      & 0.188    & 83.209    & 12.102     & 0.387     & 0.262    & 145.92    & 16.811      \\
   & 30 & 0.266     & 0.131    & 49.896    & 8.402      & 0.329     & 0.197    & 122.129   & 13.158      \\
   & 40 & 0.193     & 0.087    & 33.955    & 5.882      & 0.259     & 0.14     & 96.22     & 9.39        \\
   & 50 & \textbf{0.169}     & \textbf{0.062}    & \textbf{17.823}    & \textbf{4.285}      & \textbf{0.221}     & \textbf{0.106}    & \textbf{32.536}    & \textbf{7.178}       \\
\hline
3   & 10 & 0.274     & 0.186    & 97.894    & 12.957     & 0.42      & 0.315    & 204.197   & 19.39       \\
   & 20 & 0.143     & 0.056    & 15.844    & 4.173      & 0.235     & 0.123    & 127.802   & 8.38        \\
   & 30 & 0.102     & 0.024    & 5.618     & 1.678      & 0.139     & 0.057    & 35.617    & 3.966       \\
   & 40 & 0.055     & 0.012    & 2.549     & 0.845      & 0.095     & 0.029    & 7.355     & 2.024       \\
   & 50 & \textbf{0.04}      & \textbf{0.008}    & \textbf{1.451}     & \textbf{0.482}      & \textbf{0.049}     & \textbf{0.015}    & \textbf{3.735}     & \textbf{1.117}       \\
\hline
4   & 10 & 0.147     & 0.079    & 58.304    & 6.415      & 0.305     & 0.203    & 122.157   & 13.742      \\
   & 20 & 0.045     & 0.013    & 3.597     & 1.035      & 0.119     & 0.053    & 14.55     & 3.635       \\
   & 30 & 0.015     & 0.004    & 1.23      & 0.252      & 0.047     & 0.013    & 3.546     & 1           \\
   & 40 & 0.006     & 0.003    & 0.338     & 0.11       & 0.023     & 0.006    & 0.902     & 0.322       \\
   & 50 & \textbf{0.003}     & \textbf{0.003}    & \textbf{0.259}     & \textbf{0.076}      & \textbf{0.012}     & \textbf{0.004}    & \textbf{0.565}     & \textbf{0.199}       \\
\hline
5   & 10 & 0.081     & 0.035    & 23.325    & 3.202      & 0.208     & 0.129    & 67.093    & 9.357       \\
   & 20 & 0.012     & 0.005    & 0.879     & 0.306      & 0.052     & 0.017    & 3.522     & 1.195       \\
   & 30 & 0.003     & 0.002    & 0.256     & 0.076      & 0.019     & 0.005    & 2.238     & 0.353       \\
   & 40 & 0.001     & 0.002    & 0.208     & 0.055      & 0.004     & 0.003    & 0.309     & 0.097       \\
   & 50 &\textbf{0.0005}    & \textbf{0.002}    & \textbf{0.197}     & \textbf{0.049}      & \textbf{0.002}     & \textbf{0.003}    & \textbf{0.23}      & \textbf{0.065}       \\
\hline
6   & 10 & 0.045     & 0.017    & 5.267     & 1.562      & 0.159     & 0.08     & 28.949    & 5.847       \\
   & 20 & 0.004     & 0.003    & 0.298     & 0.1        & 0.024     & 0.008    & 1.42      & 0.539       \\
   & 30 & 0.001     & 0.002    & 0.195     & 0.048      & 0.005     & 0.002    & 0.336     & 0.115       \\
   & 40 & 0.0005    & 0.002    & 0.191     & 0.046      & 0.003     & 0.003    & 0.23      & 0.063       \\
   & 50 & \textbf{0.0004}    & \textbf{0.002}    & \textbf{0.19}      & \textbf{0.046}      & \textbf{0.001}     & \textbf{0.002}    & \textbf{0.196}     & \textbf{0.049}       \\
\bottomrule
\label{tab:dim_fs_shwi}
\end{tabular}
\end{table*}

As mentioned before, in some experiments we changed the time resolution of the HPC-DS dataset, growing it to 30 minutes, then the accuracy metrics showed in  Table~\ref{tab:dim_fs_hpc} were calculated using this new sample rate. In opposition to other datasets, AE-WMVFTS presented smaller forecast error than PCA-WMVFTS on HPC-DS dataset for the most of combinations of $K$ and $\kappa$ considering the RMSE and MAE accuracy metrics. For MAPE and SMAPE, PCA-WMVFTS was better.

\begin{table*}[!h]
\centering
\caption{PCA-WMVFTS and AE-WMVFTS model performance with different number of K-dimensions (DIM) and fuzzy sets (FS) on HPC-DS}
\begin{tabular}{ccp{1.6cm}p{1.6cm}p{1.6cm}p{2.0cm}p{1.6cm}p{1.6cm}p{1.5cm}p{1.5cm}}
\toprule
    &    & \multicolumn{4}{c}{\textbf{PCA-WMVFTS}}   & \multicolumn{4}{c}{\textbf{AE-WMVFTS}}     \\
    \toprule
DIM & FS & RMSE(Wh) & MAE(Wh) & MAP (\%) & SMAPE(\%) & RMSE(Wh) & MAE(Wh) & MAPE(\%) & SMAPE(\%)  \\
\midrule
2 & 10 & 0.549     & 0.418    & 88.727    & 24.827     & 0.507     & 0.386    & 93.654    & 25.476      \\
   & 20 & 0.511     & 0.355    & 56.347    & 17.858     & 0.383     & 0.268    & 60.252    & 19.274      \\
   & 30 & 0.469     & 0.297    & 51.887    & 15.694     & 0.34      & 0.232    & 51.57     & 17.037      \\
   & 40 & 0.41      & 0.245    & 43.124    & 13.525     & 0.324     & 0.218    & 48.976    & 16.253      \\
   & 50 & \textbf{0.366}     & \textbf{0.204}    & \textbf{37.303}    & \textbf{11.529}     & \textbf{0.304}     & \textbf{0.206}    & \textbf{46.722}    & \textbf{15.533}      \\
\hline
3   & 10 & 0.517     & 0.374    & 78.947    & 22.804     & 0.451     & 0.359    & 92.251    & 24.938      \\
   & 20 & 0.368     & 0.24     & 42.782    & 13.813     & 0.32      & 0.23     & 57.631    & 17.769      \\
   & 30 & 0.276     & 0.155    & 31.755    & 10.163     & 0.268     & 0.186    & 47.405    & 15.246      \\
   & 40 & 0.204     & 0.108    & 22.335    & 7.564      & 0.245     & 0.158    & 40.042    & 13.104      \\
   & 50 & \textbf{0.162}     & \textbf{0.079}    & \textbf{17.528}    & \textbf{5.807}      & \textbf{0.227}     & \textbf{0.137}    & \textbf{37.023}    & \textbf{11.837}      \\
\hline
4   & 10 & 0.421     & 0.309    & 70.581    & 20.532     & 0.398     & 0.325    & 90.238    & 24.065      \\
   & 20 & 0.236     & 0.147    & 28.93     & 9.689      & 0.257     & 0.186    & 51.104    & 15.894      \\
   & 30 & 0.138     & 0.071    & 15.501    & 5.438      & 0.217     & 0.143    & 41.447    & 13.013      \\
   & 40 & 0.084     & 0.039    & 8.63      & 3.147      & 0.201     & 0.116    & 33.966    & 10.828      \\
   & 50 & \textbf{0.061}     & \textbf{0.023}    & \textbf{5.24}      & \textbf{1.941}      & \textbf{0.182}     & \textbf{0.098}    & \textbf{30.187}    & \textbf{9.548}       \\
\bottomrule
\label{tab:dim_fs_hpc}
\end{tabular}
\end{table*}

Therefore, both proposed models showed  good results in both very high dimensional data, such as AEC-DS and SHWI-DS, and moderate dimensional data, such as HPC-DS, and PCA-WMVFTS is just slightly superior than AE-WMVFTS. Besides, we can improve the prediction performance of our models by just increasing a little the reduced dimensions, and $\kappa$ equal 40 or 50  seems to be the optimal number of fuzzy sets. 


\subsection{$\gamma$FTS versus Baseline models}
\label{eft_baseline}

In this subsection, we compare the forecast performance of our proposed models (PCA-WMVFTS, AE-WMVFTS and SOM-WMVFTS) with the baseline models.  For all results presented here, the number of fuzzy sets $\kappa$ was 50 and the number of $K$ dimensions was 2 for our models, and the accuracy metrics were calculated in the testing set.  

Table~\ref{tab:baselines_results_aec} presents the results of RMSE, MAE and MAPE for each baseline model on AEC-DS, as well as the accuracy metrics results for PCA-WMVFTS, AE-WMVFTS and SOM-WMVFTS proposed models.

\begin{table}[htb]
\centering
\caption{Models performance on AEC-DS dataset}
\begin{tabular}{p{3cm}p{1.5cm}p{1.5cm}p{1.5cm}}
\toprule
Method      & RMSE(Wh) & MAE(Wh) & MAPE(\%)  \\
\midrule
Persistence & 64.749    & 29.107   & 24.828     \\
SARIMAX     & 172.283   & 139.15   & 195.942    \\
PCA-SARIMAX & 158.768   & 124.465  & 170.801    \\
LSTM        & 71.329    & 42.59    & 44.748     \\
\textbf{PCA-WMVFTS}  &  \textbf{5.456}     & \textbf{2.05}     & \textbf{1.758}      \\
\textbf{AE-WMVFTS}   & \textbf{5.516}	    & \textbf{2.493}	   & \textbf{2.405}     \\
\textbf{SOM-WMVFTS}  & \textbf{18.292}     &  \textbf{7.498}   &   \textbf{4.503}    \\
\bottomrule
\label{tab:baselines_results_aec}
\end{tabular}
\end{table}

Comparing the results with those obtained by baseline models, it is clear that PCA-WMVFTS, AE-WMVFTS, and SOM-WMVFTS outperform them. Among our forecasting models, SOM-WMVFTS presented the worst performance and PCA-WMVFTS showed the smallest error, but PCA-WMVFTS is just slightly superior than AE-WMVFTS. A deep learning model such as LSTM presented a performance worse than Persistence, which was the best model among the baseline models.  

Table~\ref{tab:baselines_results_hpc} shows the accuracy for the baseline models and our proposed models on HPC-DS. We changed the time resolution of this dataset to 30 minutes, then these error metrics present the models performance for this new sample rate. 

\begin{table}[htb]
\centering
\caption{Models performance on HPC-DS dataset}
\begin{tabular}{p{3cm}p{1.5cm}p{1.5cm}p{1.5cm}}
\toprule
Method      & RMSE(kW) & MAE(kW) & MAPE(\%)  \\
\midrule
Persistence & 0.898     & 0.514    & 63.793     \\
SARIMAX     & 0.901     & 0.517    & 64.984     \\
PCA-SARIMAX & 0.901     & 0.517    & 64.944     \\
LSTM        & 0.122     & 0.088    & 17.49      \\
\textbf{PCA-WMVFTS}  & \textbf{0.366}     & \textbf{0.204}    & \textbf{37.303}     \\
\textbf{AE-WMVFTS}   & \textbf{0.304}     & \textbf{0.206}     & \textbf{46.722}      \\
\textbf{SOM-WMVFTS}  & \textbf{0.415}    & \textbf{0.256}   & \textbf{48.397}   \\
\bottomrule
\label{tab:baselines_results_hpc}
\end{tabular}
\end{table}

In contrast of AEC-DS, the LSTM showed the smallest prediction error compared to all models. In this dataset the performance of our models was worse than LSTM, however, as presented on Table~\ref{tab:dim_fs_hpc}, we can improve our model by just increasing the embedding dimension or the number of fuzzy sets, then our proposed models may be better than LSTM. Among our models, AE-WMVFTS achieved the smallest prediction error and SOM-WMVFTS the highest.  

\begin{table}[htb]
\centering
\caption{Models performance on SHWI-DS dataset}
\begin{tabular}{p{3cm}p{1.5cm}p{1.5cm}p{1.5cm}}
\toprule
Method      & RMSE(kW) & MAE(kW) & MAPE(\%)  \\
\midrule
Persistence & 0.846    & 0.468   & 251.68    \\
SARIMAX     & 1.298    & 0.719   & 284.08    \\
PCA-SARIMAX & 0.994    & 0.577   & 272.337   \\
LSTM        & 0.594    & 0.338   & 129.681   \\
\textbf{PCA-WMVFTS}  & \textbf{0.169}    & \textbf{0.062}   & \textbf{17.823}    \\
\textbf{AE-WMVFTS}   & \textbf{0.221}   & \textbf{0.106}   & \textbf{32.536}     \\
\textbf{SOM-WMVFTS}  &  \textbf{0.389} &  \textbf{0.211}   & \textbf{92.529}    \\
\bottomrule
\label{tab:baselines_results_shwi}
\end{tabular}
\end{table}

As shown on Table~\ref{tab:baselines_results_shwi}, our proposed methods is much superior than baseline models on SHWI-DS dataset. PCA-WMVFTS presented the smallest error compared to all forecasting models. SOM-WMVFTS showed again the highest error among our models, and the best baseline model was LSTM.   

Table~\ref{tab:baselines_skill_score} shows the skill score of PCA-WMVFTS, AE-WMVFTS and SOM-WMVFTS with respect to some baseline models. The accuracy metric selected was the RMSE.

\begin{table*}[!h]
\centering
\caption{Skill score of PCA-WMVFTS, AE-WMVFTS and SOM-WMVFTS with respect to baseline models}
\begin{tabular}{p{1.7cm}p{1.3cm}p{1.3cm}p{1.3cm}p{1.3cm}p{1.3cm}p{1.3cm}p{1.3cm}p{1.3cm}p{1.3cm}}
\toprule
            & \multicolumn{3}{c}{\textbf{PCA-WMVFTS}} & \multicolumn{3}{c}{\textbf{AE-WMVFTS}} & \multicolumn{3}{c}{\textbf{SOM-WMVFTS}}  \\
            \toprule
            & AEC-DS & HPC-DS & SHWI-DS               & AEC-DS & HPC-DS & SHWI-DS              & AEC-DS & HPC-DS & SHWI-DS                \\
\midrule
Persistence & 0.92   & 0.59   & 0.80                  & 0.91   & 0.66   & 0.74                 & 0.72   & 0.54   & 0.54                   \\
SARIMAX     & 0.97   & 0.59   & 0.87                  & 0.97   & 0.66   & 0.83                 & 0.89   & 0.54   & 0.70                   \\
PCA-SARIMAX & 0.97   & 0.59   & 0.83                  & 0.97   & 0.66   & 0.78                 & 0.88   & 0.54   & 0.61                   \\
LSTM        & 0.92   & -2.00  & 0.72                  & 0.92   & -1.49  & 0.63                 & 0.74   & -2.40  & 0.35                   \\
\bottomrule
\label{tab:baselines_skill_score}
\end{tabular}
\end{table*}

PCA-WMVFTS presented an improvement in RMSE greater than 90\% with respect to all baseline models on AEC-DS. In relation to SHWI-DS, the improvement is greater than 80\% with respect to persistence, SARIMAX and PCA-SARIMAX models, and PCA-WMVFTS showed an enhancement of 71\% compared to LSTM, which was the best model among baseline models. In regard to HPC-DS, PCA-WMVFTS is 66\% worse than LSTM and presented an enhancement by 59\% with respect the other baseline models.

AE-WMVFTS also showed an enhancement in RMSE greater than 90\% with respect to all baseline models on AEC-DS. In regard to SHWI-DS dataset, AE-WMVFTS presented a rise in RMSE 74\% with respect to Persistence model and it had an improvement by 83\% and 78\% with respect to SARIMAX and PCA-SARIMAX, respectively. Compared to LSTM, AE-WMVFTS has an improvement in RMSE by 63\% on SHWI-DS. In contrast, AE-WMVFTS is 59\% worse than LSTM on HPC-DS data, but it presents an improvement by 66\% with respect to the other baseline models. 

SOM-WMVFTS presented an improvement in RMSE by 89\%, 88\%, 74\% and 72\% with respect to SARIMAX, PCA-SARIMAX, Persistence and LSTM on AEC-DS, respectively. In relation to HPC-DS, SOM-WMVFTS showed a deterioration by 70\% with respect to LTSM and an enhancement by 54\% with respect to other baseline models. In regard to SHWI-DS, SOM-WMVFTS had an enhancement by 35\% compared to LSTM and it presented an increase in RMSE by 70\%, 61\% and 54\% with respect to SARIMAX, PCA-SARIMAX and Persistence, respectively.

Although our forecast models presented a negative skill score with respect to LSTM on HPC-DS, we can raise its prediction performance by just growing the number of reduced dimensions or the number of fuzzy sets. Besides, LSTM may take a considerable time to train and optimize. 

It can be seen from the results above that, compared to baseline models, PCA-WMVFTS, AE-WMVFTS and SOM-WMVFTS achieve optimal prediction performance on all datasets. The proposed models presented better results on AEC-DS and SHWI-DS, showing that they output good results in both very high dimensional data, such as AEC-DS and SHWI-DS, and moderate dimensional data, such as HPC-DS.

\subsection{$\gamma$FTS versus Competitors models}
\label{eft_competitorss}

In this subsection, we compare the performance of the proposed models with several state-of-the-art forecasting models proposed in the literature (i.e. competitors models).
It is worth noting that each one of these models used a different experimental methodology, either in terms of cross-validation methodology or train/test split. In those terms, it is counterproductive to perform a different experimental set up for each competitor model and this research used their published results as they are, to compare with our experimental results.

Table~\ref{tab:competitors_aec} shows the results of RMSE, MAE and MAPE for several state-of-the-art forecasting models proposed in the literature tested on AEC-DS with all the features and feature selection, as well as the accuracy metrics results for our proposed models. The accuracy metrics were calculated in the testing set. 

\begin{table}[htb]
\centering
\caption{Models performance on AEC-DS dataset (FS = feature selection)}
\begin{tabular}{p{3cm}p{1.5cm}p{1.5cm}p{1.5cm}}
\toprule
Method     & RMSE(Wh) & MAE(Wh) & MAPE(\%)  \\
\midrule
GBM \cite{Candanedo2017Appliances}       & 66.65    & 35.22   & 38.29     \\
GBM (FS) \cite{Candanedo2017Appliances}   & 66.21    & 35.24   & 38.65     \\
MLP \cite{Chammas2019AppliancesMLP}       & 66.29    & 29.55   & 27.96     \\
MPL (FS) \cite{Chammas2019AppliancesMLP}   & 59.84    & 27.28   & 27.09     \\
CNN-GRU \cite{Sajjad2020CNN-GRU}    & 31       & 24      & -         \\
HSBUFC   \cite{Syed2021HSBUFC}  & 5.44     & 4.45    & 2         \\
AIS-RNN  \cite{Munkhdalai2019AIS-RNN}  & 59.81    & 23.42   & 18.84     \\
\textbf{PCA-WMVFTS} & \textbf{5.456}    & \textbf{2.05}    & \textbf{1.758}     \\
\textbf{AE-WMVFTS}  & \textbf{5.516}   & \textbf{2.493}  & \textbf{2.405}    \\
\textbf{SOM-WMVFTS} & \textbf{18.292} &  \textbf{7.498}  & \textbf{11.059}   \\
\bottomrule
\label{tab:competitors_aec}
\end{tabular}
\end{table}

According to Table~\ref{tab:competitors_aec}, PCA-WMVFTS is superior than all competitors models with RMSE of 5.456 Wh, a MAE of 2.05 and a MAPE of 1.758\%. AE-WMVFTS showed a high performance compared to GBM, MPL, CNN-GRU and AIS-RNN and an accuracy error quite similar to HSBUFC. SOM-WMVFTS is inferior than HSBUFC, but it is superior than other competitors models. 

The best model among the competitors model was HSBUFC, which showed a MAE and MAPE higher than PCA-WMVFTS and a RMSE equivalent to our model. HSBUFC is just slightly superior than AE-WMVFTS taking into account RMSE and MAPE. In contrast, SOM-WMVFTS presented a performance worse than HSBUFC in all accuracy metrics. 

Table~\ref{tab:competitors_hpc} presents the accuracy for the competitors models and our models on HPC-DS. In these experiments, the time resolution of the dataset was the original one, in other words, sample rate equal one minute. PCA-WMVFTS, AE-WMVFTS and SOM-WMVFTS are inferior than HSBUFC, but our forecast models outperform the other competitors models. 

\begin{table}[htb]
\centering
\caption{Models performance on HPC-DS dataset}
\begin{tabular}{p{3cm}p{1.5cm}p{1.5cm}p{1.5cm}}
\toprule
Method     & RMSE(kW) & MAE(kW) & MAPE(\%)  \\
\midrule
CNN-LSTM \cite{Kim2019CNN-LSTM}  & 0.611    & 0.373   & 34.84     \\
M-BDLSTM  \cite{Ullah2020CNN-BI-LSTM} & 0.565    & 0.346   & 29.1      \\
CNN-GRU  \cite{Sajjad2020CNN-GRU}  & 0.47     & 0.33    & -         \\
HSBUFC   \cite{Syed2021HSBUFC}  & 0.029    & 0.022   & 3.71      \\
\textbf{PCA-WMVFTS} & \textbf{0.103}    & \textbf{0.053}   & \textbf{8.815}     \\
\textbf{AE-WMVFTS}  & \textbf{0.092}   & \textbf{0.054}   & \textbf{11.668}    \\
\textbf{SOM-WMVFTS} &  \textbf{0.095}  & \textbf{0.049}  & \textbf{9.671}         \\
\bottomrule
\label{tab:competitors_hpc}
\end{tabular}
\end{table}

Tables~\ref{tab:competitors_skill_aec} and \ref{tab:competitors_skill_hpc} show the skill score of PCA-WMVFTS, AE-WMVFTS and SOM-WMVFTS with respect to some competitors models tested on AEC-DS and HPC-DS data sets, respectively. The accuracy metric selected was the RMSE.

\begin{table}[htb]
\centering
\caption{Skill score of PCA-WMVFTS, AE-WMVFTS and SOM-WMVFTS with respect to competitors models on AEC-DS}
\begin{tabular}{p{2cm}p{1.5cm}p{1.5cm}p{1.5cm}}
\toprule
         & PCA-WMVFTS & AE-WMVFTS & SOM-WMVFTS  \\
\midrule
GBM (FS) & 0.92       & 0.92      & 0.72       \\
MPL (FS) & 0.91       & 0.91      & 0.69       \\
CNN-GRU  & 0.82       & 0.82      & 0.41       \\
HSBUFC   & 0.00      & -0.01     & -2.36      \\
AIS-RNN  & 0.91       & 0.91      & 0.69       \\
\bottomrule
\label{tab:competitors_skill_aec}
\end{tabular}
\end{table}

In relation to AEC-DS, the methods PCA-WMVFTS and AE-WMVFTS presented an improvement in RMSE greater than 90\% with respect to GBM, MLP and AIS-RNN forecast models. Both PCA-WMVFTS and AE-WMVFTS are 82\% better than CNN-GRU. Compared to HSBUFC, AE-WMVFTS had a deterioration by just 1.38\% and PCA-WMVFTS is equivalent to HSBUFC. SOM-WMVFTS had an enhancement in RMSE by 72\%, 0.69 and 0.41\% with respect to GBM, MLP, AIS-RNN and CNN-GRU, respectively. However, SOM-WMVFTS is 70\% worse than HSBUFC.  

\begin{table}
\centering
\caption{Skill score of PCA-WMVFTS, AE-WMVFTS and SOM-WMVFTS with respect to competitors models on HPC-DS}
\begin{tabular}{p{2cm}p{1.5cm}p{1.5cm}p{1.5cm}}
\toprule
         & PCA-WMVFTS & AE-WMVFTS & SOM-WMVFTS  \\
\midrule
CNN-LSTM & 0.83       & 0.85      & 0.84        \\
M-BDLSTM & 0.82       & 0.84      & 0.83        \\
CNN-GRU  & 0.78       & 0.80      & 0.80        \\
HSBUFC   & -2.55      & -2.17     & -2.28      \\
\bottomrule
\label{tab:competitors_skill_hpc}
\end{tabular}
\end{table}

In regard to HPC-DS dataset, our three proposed models showed an improvement in RMSE with respect to the CNN-LSTM, M-BDLSTM and CNN-GRU models. AE-WMVFTS and SOM-WMVFTS had an enhancement greater than 80\% compared to these competitors models. PCA-WMVFTS is 78\% better than CNN-GRU and presentd an enhancement greater than 80\% with respect to  CNN-LSTM and M-BDLSTM. In contrast, compared to HSBUFC, our approach with two-dimensional embedding and 50 fuzzy sets showed a deterioration around 70\%.

Although our models showed a negative skill score with respect to HSBUFC, we can raise their prediction performance by just growing the number of reduced dimensions or the number of fuzzy sets. Additionally, HSBUFC is a deep leaning model that may take a considerable time to train and optimize, while FTS models are very fast to train. 

It can be concluded that our three proposed methods with 2 reduced dimensions and 50 fuzzy sets significantly outperformed several competitors models, outputting a consistent and accurate prediction, since the methods presented equally good results in all the data sets that were used, reinforcing the model's capacity to make excellent predictions for different data.

\subsection{Analysis and Discussion}

This section aims to discuss the accuracy, parsimony, computational cost and explainability of $\gamma$FTS models considering the results of the experiments.

As a rule of thumb, increasing the embedding dimension and the number of fuzzy sets increase the accuracy of $\gamma$FTS models, except for the SOM function. However, there is a clear trade-off:  increasing these hyperparameters improves accuracy but decreases the parsimony and explainability. 

The parsimony, or complexity, is intended to be minimized once it means the number of parameters or rules of the model. As the model complexity increases also its computational cost and training time increase. This is relevant in the domain of IoE applications, since these models can be embedded in edge devices with low computational power. $\gamma$FTS models are less parsimonious than SARIMAX but way more parsimonious, for instance, than deep learning models such as LSTM and CNN.  

It is also worth observing that the broad class of FTS models are data-driven and white-box models that are easy to explain and audit, a feature that gained high importance in recent years. $\gamma$FTS models have their explainability affected by two factors: the embedding function and the number of fuzzy sets. The most impacting factor is the number of fuzzy sets which automatically increases the number of rules and makes the model more complex. The embedding function is not a complex issue per se, since the presented functions are all invertible, and the values of the fuzzy sets can be converted back to values of the original data space.

Future investigations may also focus on the properties of the feature spaces of each embedding function. It is important to note that the PCA, which is a linear embedding, in general, performed better than AE and SOM, which are nonlinear embeddings. The less accurate embedding in general was the SOM, and a possible hypothesis for this is due to their discrete embedding space, but it is exactly this property that makes it more explainable due to the easiness to identify features in the embedded space.

In this work, the proposed approach aims to address the smart building energy consumption forecasting problem. The embedding algorithm is used to extract the main information that better represent the content of multivariate time series for the subsequent energy forecasting task. PCA, autoencoder and SOM algorithms can be used to identify the most relevant information on the smart buildings datasets based on available historical data. 


\section{Conclusions and Future Works}
\label{conclusions}

In this work, we investigated the possible benefits provided by a method that combines embedding transformation and fuzzy time series forecasting approach for tackling the high-dimensional time series forecasting problem. We proposed a new methodology (namely $\gamma$FTS) for handling high-dimensional data, applying data embedding transformation and FTS model in the low-dimensional learned representation space.

Since each variable $\mathcal{V}$ depends not only on its historic values but also of several contemporaneous variables, it is mandatory to select a set of suitable variables as input. In this sense, the embedding techniques allow us to extract and exploit a new feature space that better represents the inherent complexity of multivariate time series, also mitigating collinearity phenomena and catching latent interactions among features. 

High dimensional time series are a big challenge for forecasting methods. Statistical models can suffer from the curse of dimensionality and  multi-collinearity, and the existing fuzzy time series models can be unfeasible if all features were used to train the model. On the positive side, the flexibility, simplicity, readability and accuracy of FTS methods make this approach interesting for many IoE applications. 

The proposed approach aimed to address the energy consumption forecasting problem in smarts buildings. The PCA, autoencoder and SOM algorithms were used to extract new feature space that better represents the content of energy consumption multivariate time series for the subsequent forecasting task. The embedding methods allow us to extract the relevant information that supports the target variable forecasting. 

Computational experiments were performed to assess the accuracy of $\gamma$FTS against some baseline models (persistence, SARIMAX, PCA-SARIMAX and LSTM) and state-of-the-art machine learning and deep learning models recently proposed in the literature.

Our experimental evaluation showed that, compared to other energy consumption forecasting methods, $\gamma$FTS achieves the best prediction performance on smart building energy consumption problem. Therefore, our approach has a great value in smart building and IoE applications, and can help homeowners reduce their power consumption and provide better energy-saving strategies. 

Besides, the proposed methodology generates forecasting models readable and explainable and their accuracy are controlled basically by two parameters: the partitioning of the target variable (number of fuzzy sets) and the embedding dimension $K$. As the number of reduced dimensions and fuzzy sets increase, the forecast error decreases, then we can improve our approach by just increasing these two hyperparameters when needed. 

As future work, we intend to extend the proposed methods to handle multivariate non-stationary time series and the implementation of a MIMO (multiple input multiple output) and many step-ahead forecasting approach.  

\bibliography{references}

\begin{thebibliography}{10}
\expandafter\ifx\csname url\endcsname\relax
  \def\url#1{\texttt{#1}}\fi
\expandafter\ifx\csname urlprefix\endcsname\relax\def\urlprefix{URL }\fi
\expandafter\ifx\csname href\endcsname\relax
  \def\href#1#2{#2} \def\path#1{#1}\fi

\bibitem{Union2015IoTReport}
I.~T. Union, Itu internet report 2005: The internet of things (2005).

\bibitem{Gubbi2013IoTSurvey}
J.~Gubbi, R.~Buyya, S.~Marusic, M.~Palaniswami,
  \href{https://doi.org/10.1016/j.future.2013.01.010}{Internet of things (iot):
  {A} vision, architectural elements, and future directions}, Future Gener.
  Comput. Syst. 29~(7) (2013) 1645--1660.
\newblock \href {https://doi.org/10.1016/j.future.2013.01.010}
  {\path{doi:10.1016/j.future.2013.01.010}}.
\newline\urlprefix\url{https://doi.org/10.1016/j.future.2013.01.010}

\bibitem{Miorandi2012IoTSurvey}
D.~Miorandi, S.~Sicari, F.~{De Pellegrini}, I.~Chlamtac,
  \href{https://www.sciencedirect.com/science/article/pii/S1570870512000674}{Internet
  of things: Vision, applications and research challenges}, Ad Hoc Networks
  10~(7) (2012) 1497--1516.
\newblock \href {https://doi.org/https://doi.org/10.1016/j.adhoc.2012.02.016}
  {\path{doi:https://doi.org/10.1016/j.adhoc.2012.02.016}}.
\newline\urlprefix\url{https://www.sciencedirect.com/science/article/pii/S1570870512000674}

\bibitem{Reka2018FutureEffectualIoT}
S.~Reka, Future effectual role of energy delivery: A comprehensive review of
  internet of things and smart grid, Renewable and Sustainable Energy Reviews
  91 (04 2018).
\newblock \href {https://doi.org/10.1016/j.rser.2018.03.089}
  {\path{doi:10.1016/j.rser.2018.03.089}}.

\bibitem{Yasir2020IoE}
Y.~Shahzad, H.~Javed, H.~Farman, J.~Ahmad, B.~Jan, M.~Zubair,
  \href{https://doi.org/10.1016/j.compeleceng.2020.106739}{Internet of energy:
  Opportunities, applications, architectures and challenges in smart
  industries}, Comput. Electr. Eng. 86 (2020) 106739.
\newblock \href {https://doi.org/10.1016/j.compeleceng.2020.106739}
  {\path{doi:10.1016/j.compeleceng.2020.106739}}.
\newline\urlprefix\url{https://doi.org/10.1016/j.compeleceng.2020.106739}

\bibitem{Michele2016IoTSurvey}
M.~Nitti, V.~Pilloni, G.~Colistra, L.~Atzori,
  \href{https://doi.org/10.1109/COMST.2015.2498304}{The virtual object as a
  major element of the internet of things: {A} survey}, {IEEE} Commun. Surv.
  Tutorials 18~(2) (2016) 1228--1240.
\newblock \href {https://doi.org/10.1109/COMST.2015.2498304}
  {\path{doi:10.1109/COMST.2015.2498304}}.
\newline\urlprefix\url{https://doi.org/10.1109/COMST.2015.2498304}

\bibitem{Manar2015IoE}
M.~Jaradat, M.~H.~A. Jarrah, A.~Bousselham, Y.~Jararweh, M.~Al{-}Ayyoub,
  \href{https://doi.org/10.1016/j.procs.2015.07.250}{The internet of energy:
  Smart sensor networks and big data management for smart grid}, in: The 10th
  International Conference on Future Networks and Communications {(FNC} 2015) /
  The 12th International Conference on Mobile Systems and Pervasive Computing
  (MobiSPC 2015) / Affiliated Workshops, August 17-20, 2015, Belfort, France,
  Vol.~56 of Procedia Computer Science, Elsevier, 2015, pp. 592--597.
\newblock \href {https://doi.org/10.1016/j.procs.2015.07.250}
  {\path{doi:10.1016/j.procs.2015.07.250}}.
\newline\urlprefix\url{https://doi.org/10.1016/j.procs.2015.07.250}

\bibitem{Shahinzadeh2019IoE}
H.~Shahinzadeh, J.~Moradi, G.~B.~Gharehpetian, H.~Nafisi, M.~Abedi, Internet of
  energy (ioe) in smart power systems, 2019, pp. 627--636.
\newblock \href {https://doi.org/10.1109/KBEI.2019.8735086}
  {\path{doi:10.1109/KBEI.2019.8735086}}.

\bibitem{Jaradat2015IotSmartGrid}
M.~Jaradat, M.~Jarrah, A.~Bousselham, Y.~Jararweh, M.~Al-Ayyoub,
  \href{https://doi.org/10.1016%2Fj.procs.2015.07.250}{The internet of energy:
  Smart sensor networks and big data management for smart grid}, Procedia
  Computer Science 56 (2015) 592--597.
\newblock \href {https://doi.org/10.1016/j.procs.2015.07.250}
  {\path{doi:10.1016/j.procs.2015.07.250}}.
\newline\urlprefix\url{https://doi.org/10.1016%2Fj.procs.2015.07.250}

\bibitem{Lobaccaro2016SurveySmartHomes}
G.~Lobaccaro, S.~Carlucci, E.~Löfström,
  \href{https://www.mdpi.com/1996-1073/9/5/348}{A review of systems and
  technologies for smart homes and smart grids}, Energies 9~(5) (2016).
\newblock \href {https://doi.org/10.3390/en9050348}
  {\path{doi:10.3390/en9050348}}.
\newline\urlprefix\url{https://www.mdpi.com/1996-1073/9/5/348}

\bibitem{Manic2016IntelligentBuildings}
M.~Manic, K.~Amarasinghe, J.~J. Rodriguez-Andina, C.~Rieger, Intelligent
  buildings of the future: Cyberaware, deep learning powered, and human
  interacting, IEEE Industrial Electronics Magazine 10~(4) (2016) 32--49.
\newblock \href {https://doi.org/10.1109/MIE.2016.2615575}
  {\path{doi:10.1109/MIE.2016.2615575}}.

\bibitem{Hammami2020ANNOnlineSmartGrids}
Z.~Hammami, M.~{Sayed Mouchaweh}, W.~Mouelhi, L.~B. Said,
  \href{https://doi.org/10.1007/s10462-020-09844-3}{Neural networks for online
  learning of non-stationary data streams: a review and application for smart
  grids flexibility improvement}, Artif. Intell. Rev. 53~(8) (2020) 6111--6154.
\newblock \href {https://doi.org/10.1007/s10462-020-09844-3}
  {\path{doi:10.1007/s10462-020-09844-3}}.
\newline\urlprefix\url{https://doi.org/10.1007/s10462-020-09844-3}

\bibitem{Mohammadi2018SurveyDPIoTBigData}
M.~Mohammadi, A.~Al-Fuqaha, S.~Sorour, M.~Guizani, Deep learning for iot big
  data and streaming analytics: A survey, IEEE Communications Surveys Tutorials
  20~(4) (2018) 2923--2960.
\newblock \href {https://doi.org/10.1109/COMST.2018.2844341}
  {\path{doi:10.1109/COMST.2018.2844341}}.

\bibitem{Guimaraes2020HyperOtimFTS}
P.~C. de~Lima~Silva, P.~de~Oliveira~e Lucas, H.~J. Sadaei, F.~G.
  Guimar{\~{a}}es, \href{https://doi.org/10.1109/TNSM.2020.2980289}{Distributed
  evolutionary hyperparameter optimization for fuzzy time series}, {IEEE}
  Trans. Netw. Serv. Manag. 17~(3) (2020) 1309--1321.
\newblock \href {https://doi.org/10.1109/TNSM.2020.2980289}
  {\path{doi:10.1109/TNSM.2020.2980289}}.
\newline\urlprefix\url{https://doi.org/10.1109/TNSM.2020.2980289}

\bibitem{Severiano2020NSFTS}
P.~C. de~Lima~Silva, C.~A.~S. Junior, M.~A. Alves, R.~Silva, M.~Weiss{-}Cohen,
  F.~G. Guimar{\~{a}}es,
  \href{https://doi.org/10.1016/j.asoc.2020.106825}{Forecasting in
  non-stationary environments with fuzzy time series}, Applied Soft Computing
  97~(Part {B}) (2020) 106825.
\newblock \href {https://doi.org/10.1016/j.asoc.2020.106825}
  {\path{doi:10.1016/j.asoc.2020.106825}}.
\newline\urlprefix\url{https://doi.org/10.1016/j.asoc.2020.106825}

\bibitem{Zheng2015WSN}
A.~Zheng, J.~Jamelipour, Wireless Sensor Networks: A Networking Perspective,
  Wiley-{IEEE} Press, 2009.

\bibitem{Vieira2003WSNSurvey}
M.~A.~M. Vieira, J.~Coelho, Claudionor~N., J.~da~Silva, Diógenes.Cecílio.,
  J.~da~Mata, Survey on wireless sensor network devices, in: {IEEE} Conference
  on Emerging Technologies and Factory Automation (ETFA), 2003, Vol.~1, 2003,
  pp. 537 -- 544 vol.1.
\newblock \href {https://doi.org/10.1109/ETFA.2003.1247753}
  {\path{doi:10.1109/ETFA.2003.1247753}}.

\bibitem{Mocanu2016DPBuildingEnergyConsumption}
E.~Mocanu, P.~H. Nguyen, M.~Gibescu, W.~L. Kling,
  \href{https://www.sciencedirect.com/science/article/pii/S2352467716000163}{Deep
  learning for estimating building energy consumption}, Sustainable Energy,
  Grids and Networks 6 (2016) 91--99.
\newblock \href {https://doi.org/https://doi.org/10.1016/j.segan.2016.02.005}
  {\path{doi:https://doi.org/10.1016/j.segan.2016.02.005}}.
\newline\urlprefix\url{https://www.sciencedirect.com/science/article/pii/S2352467716000163}

\bibitem{Candanedo2017Appliances}
L.~M. Candanedo, V.~Feldheim, D.~Deramaix,
  \href{https://www.sciencedirect.com/science/article/pii/S0378778816308970}{Data
  driven prediction models of energy use of appliances in a low-energy house},
  Energy and Buildings 140 (2017) 81--97.
\newblock \href {https://doi.org/https://doi.org/10.1016/j.enbuild.2017.01.083}
  {\path{doi:https://doi.org/10.1016/j.enbuild.2017.01.083}}.
\newline\urlprefix\url{https://www.sciencedirect.com/science/article/pii/S0378778816308970}

\bibitem{Chammas2019AppliancesMLP}
M.~Chammas, A.~Makhoul, J.~Demerjian,
  \href{https://doi.org/10.1016/j.compeleceng.2019.04.002}{An efficient data
  model for energy prediction using wireless sensors}, Comput. Electr. Eng. 76
  (2019) 249--257.
\newblock \href {https://doi.org/10.1016/j.compeleceng.2019.04.002}
  {\path{doi:10.1016/j.compeleceng.2019.04.002}}.
\newline\urlprefix\url{https://doi.org/10.1016/j.compeleceng.2019.04.002}

\bibitem{Syed2021HSBUFC}
D.~Syed, H.~Abu-Rub, A.~Ghrayeb, S.~S. Refaat, Household-level energy
  forecasting in smart buildings using a novel hybrid deep learning model, IEEE
  Access 9 (2021) 33498--33511.
\newblock \href {https://doi.org/10.1109/ACCESS.2021.3061370}
  {\path{doi:10.1109/ACCESS.2021.3061370}}.

\bibitem{Kim2019CNN-LSTM}
T.-Y. Kim, S.-B. Cho,
  \href{https://www.sciencedirect.com/science/article/pii/S0360544219311223}{Predicting
  residential energy consumption using cnn-lstm neural networks}, Energy 182
  (2019) 72--81.
\newblock \href {https://doi.org/https://doi.org/10.1016/j.energy.2019.05.230}
  {\path{doi:https://doi.org/10.1016/j.energy.2019.05.230}}.
\newline\urlprefix\url{https://www.sciencedirect.com/science/article/pii/S0360544219311223}

\bibitem{Sajjad2020CNN-GRU}
M.~Sajjad, Z.~A. Khan, A.~Ullah, T.~Hussain, W.~Ullah, M.~Y. Lee, S.~W. Baik, A
  novel cnn-gru-based hybrid approach for short-term residential load
  forecasting, IEEE Access 8 (2020) 143759--143768.
\newblock \href {https://doi.org/10.1109/ACCESS.2020.3009537}
  {\path{doi:10.1109/ACCESS.2020.3009537}}.

\bibitem{Ullah2020CNN-BI-LSTM}
F.~U.~M. Ullah, A.~Ullah, I.~U. Haq, S.~Rho, S.~W. Baik, Short-term prediction
  of residential power energy consumption via cnn and multi-layer
  bi-directional lstm networks, IEEE Access 8 (2020) 123369--123380.
\newblock \href {https://doi.org/10.1109/ACCESS.2019.2963045}
  {\path{doi:10.1109/ACCESS.2019.2963045}}.

\bibitem{Munkhdalai2019AIS-RNN}
L.~Munkhdalai, T.~Munkhdalai, K.~H. Park, T.~Amarbayasgalan, E.~Batbaatar,
  H.~W. Park, K.~H. Ryu, An end-to-end adaptive input selection with dynamic
  weights for forecasting multivariate time series, IEEE Access 7 (2019)
  99099--99114.
\newblock \href {https://doi.org/10.1109/ACCESS.2019.2930069}
  {\path{doi:10.1109/ACCESS.2019.2930069}}.

\bibitem{Parhizkar2021PCAEnergyConsuption}
T.~Parhizkar, E.~Rafieipour, A.~Parhizkar,
  \href{https://www.sciencedirect.com/science/article/pii/S0959652620339111}{Evaluation
  and improvement of energy consumption prediction models using principal
  component analysis based feature reduction}, Journal of Cleaner Production
  279 (2021) 123866.
\newblock \href {https://doi.org/https://doi.org/10.1016/j.jclepro.2020.123866}
  {\path{doi:https://doi.org/10.1016/j.jclepro.2020.123866}}.
\newline\urlprefix\url{https://www.sciencedirect.com/science/article/pii/S0959652620339111}

\bibitem{Song1993FTS}
Q.~Song, B.~S. Chissom,
  \href{https://www.sciencedirect.com/science/article/pii/016501149390372O}{Fuzzy
  time series and its models}, Fuzzy Sets and Systems 54~(3) (1993) 269--277.
\newblock \href {https://doi.org/https://doi.org/10.1016/0165-0114(93)90372-O}
  {\path{doi:https://doi.org/10.1016/0165-0114(93)90372-O}}.
\newline\urlprefix\url{https://www.sciencedirect.com/science/article/pii/016501149390372O}

\bibitem{Pearson1901PCA}
K.~P. F.R.S., Liii. on lines and planes of closest fit to systems of points in
  space, The London, Edinburgh, and Dublin Philosophical Magazine and Journal
  of Science 2~(11) (1901) 559--572.
\newblock \href {https://doi.org/10.1080/14786440109462720}
  {\path{doi:10.1080/14786440109462720}}.

\bibitem{Rumelhart1986Autoencoders}
D.~E. Rumelhart, G.~E. Hinton, R.~J. Williams, Learning internal
  representations by error propagation, in: D.~E. Rumelhart, J.~L. Mcclelland
  (Eds.), Parallel Distributed Processing: Explorations in the Microstructure
  of Cognition, {V}olume 1: {F}oundations, MIT Press, Cambridge, MA, 1986, pp.
  318--362.

\bibitem{Kohonen2001SOM}
T.~Kohonen, \href{https://doi.org/10.1007/978-3-642-56927-2}{Self-Organizing
  Maps, Third Edition}, Springer Series in Information Sciences, Springer,
  2001.
\newblock \href {https://doi.org/10.1007/978-3-642-56927-2}
  {\path{doi:10.1007/978-3-642-56927-2}}.
\newline\urlprefix\url{https://doi.org/10.1007/978-3-642-56927-2}

\bibitem{Santos2021SOM-FTS}
M.~C. dos Santos, F.~G. Guimarães, P.~C. d.~L. Silva, High-dimensional
  multivariate time series forecasting using self-organizing maps and fuzzy
  time series, in: 2021 IEEE International Conference on Fuzzy Systems
  (FUZZ-IEEE), 2021, pp. 1--6.
\newblock \href {https://doi.org/10.1109/FUZZ45933.2021.9494496}
  {\path{doi:10.1109/FUZZ45933.2021.9494496}}.

\bibitem{Bitencourt2021ENSFTS}
H.~V. Bitencourt, F.~G. Guimar{\~{a}}es,
  \href{https://arxiv.org/abs/2107.09785}{High-dimensional multivariate time
  series forecasting in iot applications using embedding non-stationary fuzzy
  time series}, CoRR abs/2107.09785 (2021).
\newblock \href {http://arxiv.org/abs/2107.09785} {\path{arXiv:2107.09785}}.
\newline\urlprefix\url{https://arxiv.org/abs/2107.09785}

\bibitem{Gensler2016AE-LSTMPV}
A.~Gensler, J.~Henze, B.~Sick, N.~Raabe, Deep learning for solar power
  forecasting - an approach using autoencoder and lstm neural networks, in:
  2016 IEEE International Conference on Systems, Man, and Cybernetics (SMC),
  2016, pp. 002858--002865.
\newblock \href {https://doi.org/10.1109/SMC.2016.7844673}
  {\path{doi:10.1109/SMC.2016.7844673}}.

\bibitem{e2019distributed}
P.~C. de~Lima~{Silva}, e~Patrícia de Oliveira~e {Lucas}, F.~G. {Guimarães},
  \href{https://doi.org/10.1007/978-3-030-19223-5_4}{A distributed algorithm
  for scalable fuzzy time series}, in: International Conference on Green,
  Pervasive, and Cloud Computing, Springer, 2019, pp. 42--56.
\newblock \href {https://doi.org/10.1007/978-3-030-19223-5\_4}
  {\path{doi:10.1007/978-3-030-19223-5\_4}}.
\newline\urlprefix\url{https://doi.org/10.1007/978-3-030-19223-5_4}

\bibitem{Dua2019UCIDatasets}
D.~Dua, C.~Graff, \href{http://archive.ics.uci.edu/ml}{{UCI} machine learning
  repository} (2017).
\newline\urlprefix\url{http://archive.ics.uci.edu/ml}

\bibitem{KaggleDataset}
Kaggle, Smart home dataset with weather information,
  \url{https://www.kaggle.com/taranvee/smart-home-dataset-with-weather-information},
  accessed on 28 Ago 2021 (2021).

\bibitem{Cyril2017MLPV}
C.~Voyant, G.~Notton, S.~Kalogirou, M.-L. Nivet, C.~Paoli, F.~Motte,
  A.~Fouilloy,
  \href{http://www.sciencedirect.com/science/article/pii/S0960148116311648}{Machine
  learning methods for solar radiation forecasting: A review}, Renewable Energy
  105 (2017) 569 -- 582.
\newblock \href {https://doi.org/https://doi.org/10.1016/j.renene.2016.12.095}
  {\path{doi:https://doi.org/10.1016/j.renene.2016.12.095}}.
\newline\urlprefix\url{http://www.sciencedirect.com/science/article/pii/S0960148116311648}

\bibitem{pmdarima}
T.~G. Smith, et~al., \href{http://www.alkaline-ml.com/pmdarima}{{pmdarima}:
  Arima estimators for {Python}}, [Online; accessed <today>] (2017--).
\newline\urlprefix\url{http://www.alkaline-ml.com/pmdarima}

\bibitem{Hochreiter1997LSTM}
S.~Hochreiter, J.~Schmidhuber, Long short-term memory, Neural computation 9~(8)
  (1997) 1735--1780.

\bibitem{Bergstra2013Hyperopt}
J.~Bergstra, D.~Yamins, D.~Cox, Making a science of model search:
  Hyperparameter optimization in hundreds of dimensions for vision
  architectures (2013) 115--123.

\bibitem{pyFTS}
P.~C. d.~L. Silva, pyfts : Fuzzy time series for python (03 2016).
\newblock \href {https://doi.org/10.5281/zenodo.597359}
  {\path{doi:10.5281/zenodo.597359}}.

\bibitem{scikit-learn}
F.~Pedregosa, G.~Varoquaux, A.~Gramfort, V.~Michel, B.~Thirion, O.~Grisel,
  M.~Blondel, P.~Prettenhofer, R.~Weiss, V.~Dubourg, J.~Vanderplas, A.~Passos,
  D.~Cournapeau, M.~Brucher, M.~Perrot, E.~Duchesnay, Scikit-learn: Machine
  learning in {P}ython, Journal of Machine Learning Research 12 (2011)
  2825--2830.

\bibitem{keras}
F.~Chollet, et~al., Keras, \url{https://keras.io} (2015).

\bibitem{tensorflow}
M.~Abadi, A.~Agarwal, P.~Barham, E.~Brevdo, Z.~Chen, C.~Citro, G.~S. Corrado,
  A.~Davis, J.~Dean, M.~Devin, S.~Ghemawat, I.~Goodfellow, A.~Harp, G.~Irving,
  M.~Isard, Y.~Jia, R.~Jozefowicz, L.~Kaiser, M.~Kudlur, J.~Levenberg,
  D.~Man\'{e}, R.~Monga, S.~Moore, D.~Murray, C.~Olah, M.~Schuster, J.~Shlens,
  B.~Steiner, I.~Sutskever, K.~Talwar, P.~Tucker, V.~Vanhoucke, V.~Vasudevan,
  F.~Vi\'{e}gas, O.~Vinyals, P.~Warden, M.~Wattenberg, M.~Wicke, Y.~Yu,
  X.~Zheng, \href{https://www.tensorflow.org/}{{TensorFlow}: Large-scale
  machine learning on heterogeneous systems}, software available from
  tensorflow.org (2015).
\newline\urlprefix\url{https://www.tensorflow.org/}

\bibitem{statsmodels}
S.~Seabold, J.~Perktold, statsmodels: Econometric and statistical modeling with
  python, in: 9th Python in Science Conference, 2010.

\end{thebibliography}

\end{document}